\documentclass[11pt, a4paper, logo, copyright]{googledeepmind}

\pdftrailerid{redacted}

\makeatletter
\renewcommand\bibentry[1]{\nocite{#1}{\frenchspacing\@nameuse{BR@r@#1\@extra@b@citeb}}}
\makeatother

\usepackage{kantlipsum, lipsum}
\usepackage{dsfont}
\usepackage{gdm-colors}
\usepackage[utf8]{inputenc}   
\usepackage{newunicodechar}   
\usepackage{amssymb}          
\usepackage{arydshln}
\usepackage{wasysym,marvosym}
 
\usepackage[authoryear, sort&compress, round]{natbib}

\usepackage{bbding}
\usepackage[utf8]{inputenc} 
\usepackage[T1]{fontenc}    
\usepackage{hyperref}       
\usepackage{url}            
\usepackage{booktabs}       
\usepackage{nicefrac}       
\usepackage{microtype}      
\usepackage{amsmath}
\usepackage{graphicx}
\usepackage{tablefootnote}
\usepackage{multicol}
\usepackage[nameinlink]{cleveref}
\usepackage{bbm}
\usepackage{multirow}
\usepackage{soul}
\usepackage{float}
\usepackage{wrapfig}
\usepackage{blindtext}
\usepackage{tablefootnote}
\usepackage{amsfonts}
\usepackage[flushleft]{threeparttable}
\usepackage{colortbl}
\usepackage{mathtools,amssymb}
\usepackage{bm}
\usepackage{makecell}
\usepackage{caption}
\usepackage{capt-of}
\usepackage{array}
\usepackage{calc}      
\usepackage{caption}   
\usepackage{subcaption}  
\usepackage{xcolor,colortbl}
\usepackage[bottom]{footmisc}

\usepackage{tcolorbox}

\definecolor{mydeepgreen}{RGB}{0, 100, 0} 

\newcommand{\myred}[1]{\textcolor{red}{#1}}

\usepackage{xspace}

\newcommand{\myorange}[1]{\textcolor{dmorange500}{#1}}

\newcommand{\cdashlinerow}[2]{%
  \cdashline{#1}%
  \noalign{\global\let\CT@row@color\relax\vskip0pt}%
  \rowcolor{#2}%
}

\newcommand{\eat}[1]{}

\crefformat{section}{\S#2#1#3}

\graphicspath{{figures/}}

\title{Kelix Technical Report: \\ Closing the Understanding Gap of Discrete Tokens in Unified Multimodal Models}

\renewcommand{\today}{}

\author{\large OneRec Team, Kuaishou Technology}

\begin{abstract}
    Autoregressive large language models (LLMs) scale well by expressing diverse tasks as sequences of discrete natural-language tokens and training with next-token prediction, which unifies comprehension and generation under self-supervision. Extending this paradigm to multimodal data requires a shared, discrete representation across modalities.
    However, most vision-language models (VLMs) still rely on a hybrid interface: discrete text tokens paired with continuous Vision Transformer (ViT) features. Because supervision is largely text-driven, these models are often biased toward understanding and cannot fully leverage large-scale self-supervised learning on non-text data.
    Recent work has explored discrete visual tokenization to enable fully autoregressive multimodal modeling, showing promising progress toward unified understanding and generation. Yet existing discrete vision tokens frequently lose information due to limited code capacity, resulting in noticeably weaker understanding than continuous-feature VLMs.
    We present \myorange{Kelix}, a fully discrete autoregressive unified model that closes the understanding gap between discrete and continuous visual representations. At its core is a product-quantization-based vision tokenization mechanism that increases the information capacity of discrete codes, enabling discrete-token models to achieve understanding performance comparable to their continuous-feature counterparts. Extensive experiments demonstrate state-of-the-art results on most multimodal understanding and image generation benchmarks among comparable-scale unified models. In particular, Kelix achieves significant improvements over prior discrete approaches on understanding tasks, especially on text-rich benchmarks, where discrete methods have been weakest, surpassing the previous best discrete model by over 23\% on OCRBench and closely approaching continuous baselines. This demonstrates that discrete representations are no longer the bottleneck of multimodal understanding, offering new insights for the future development of discrete unified models.
\end{abstract}

\begin{document}

\maketitle

\begin{figure}[h!]
    \centering
    \includegraphics[width=1\textwidth]{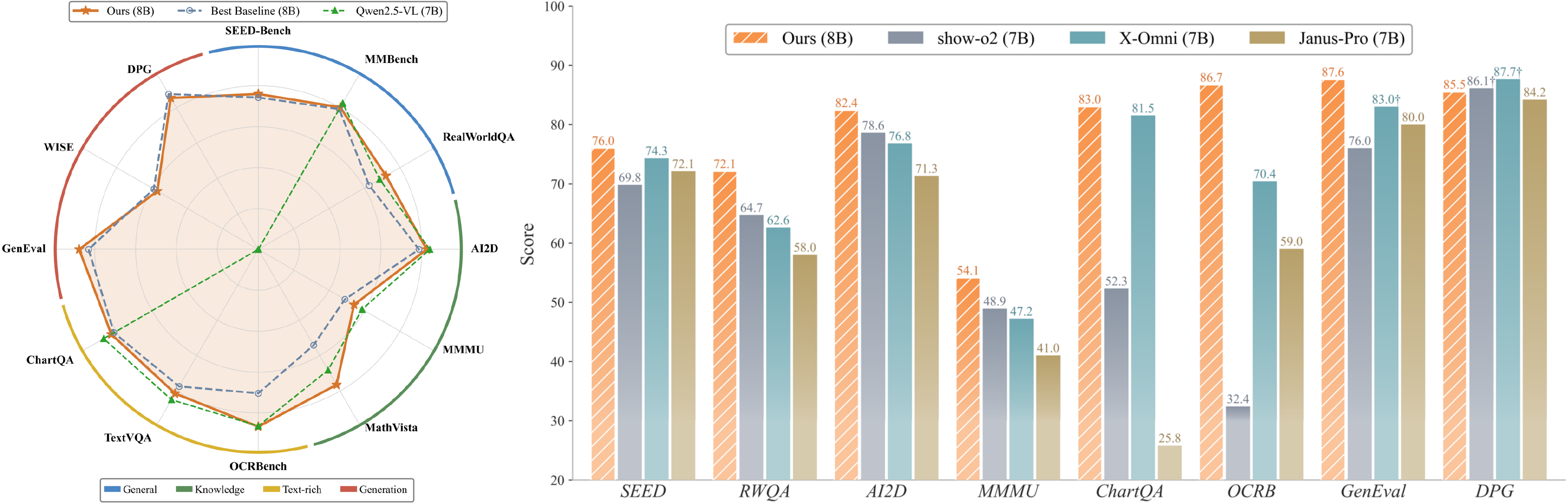}
      \caption{Benchmark performance of Kelix. Compared with prior discrete unified models, Kelix significantly improves understanding, even matching continuous understanding-only VLMs, while achieving competitive image generation results with current SOTA.}
      \label{rader}
\end{figure}

\section{Introduction}

Autoregressive large language models (LLMs) owe much of their scaling success to a unified discrete representation: by expressing diverse tasks as sequences of tokens and training with next-token prediction, a single objective serves both comprehension and generation~\citep{radford2018improving,radford2019language,brown2020language,achiam2023gpt,agarwal2025gpt,kaplan2020scaling}. Extending this paradigm to multimodal data, where vast non-text corpora remain underexploited, promises a new scaling frontier.
Yet mainstream vision-language models (VLMs) still rely on a hybrid interface: discrete text tokens paired with continuous Vision Transformer (ViT) features~\citep{liu2023visual,dosovitskiy2020image}. Because supervision is largely text-driven, vision often collapses into a conditioning signal, yielding understanding-only models that cannot fully leverage self-supervised learning on non-text data~\citep{yao2024minicpm,chen2024internvl,team2025kimi}.
A natural solution is to align vision to the discrete token interface of LLMs via discrete image tokenizers, enabling fully autoregressive multimodal modeling that unifies understanding and generation. However, existing discrete-token approaches suffer from limited information capacity: a single discrete code represents far less information than the continuous embedding it replaces, resulting in noticeably weaker multimodal understanding than continuous-feature VLMs.

In this paper, we propose \textbf{Kelix}\footnote{The name \emph{Kelix} is a portmanteau of \emph{K}uaishou and h\emph{elix}: just as the DNA double helix encodes the full complexity of life using only four discrete nucleotide bases (A, T, C, G), Kelix encodes rich visual semantics using discrete tokens.}, a fully discrete, LLM-centric unified model that closes this understanding gap. At its core is a product-quantization-based vision tokenization scheme that expands the coding space of discrete visual representations, substantially reducing information loss. The tokenizer is learned end-to-end with the LLM, allowing quantization to adapt to the downstream objective. We further adopt a next-block prediction objective that models grouped discrete token blocks, mitigating context-length explosion while preserving multimodal information. Extensive experiments demonstrate state-of-the-art results on major understanding and generation benchmarks among comparable-scale unified models\footnote{Our training scripts at \url{https://github.com/Kuaishou-OneRec/Kelix} and checkpoints at \url{https://huggingface.co/collections/OpenOneRec/kelix}.}.

\section{Design Philosophy}
\label{sec2_Design}

Before describing the architecture in detail, we outline the design principles that guided our choices. Unified multimodal modeling has not yet converged on a single dominant design. Rather than exploring the full space empirically, we take \emph{representation design} as the entry point: how should input/output token representations be defined so that modalities can be aligned and fused within a single LLM-centric framework?

We factor representation design into two largely orthogonal dimensions:
\begin{itemize}
    \item \myorange{\textbf{Representation form}}: discrete vs.\ continuous.
    \item \myorange{\textbf{Semantic level}}: high-level semantics vs.\ low-level semantics.
\end{itemize}
Our choices are guided by two principles: (1)~\textbf{strong scaling potential}---the representation should support efficient scaling via high sample efficiency, high MFU, and a unified space for cross-modal transfer; and (2)~\textbf{maximum compatibility with the LLM ecosystem}---reusing open foundation models, training frameworks, and deployment toolchains.

\textbf{Semantic level.} We prefer \emph{high-level semantics} as the unified representation. High-level tokens are more compressible and align naturally with text, benefiting sample efficiency and cross-modal transfer. To maximize MFU, we decouple the system into a modular \emph{Tokenizer--LLM--Detokenizer} pipeline: the LLM backbone performs scalable sequence modeling in the semantic space, while modality-specific encoding and decoding are delegated to the Tokenizer and Detokenizer, which interact with the LLM only through the unified representation interface.

\textbf{Representation form.} We favor \emph{discrete} tokens as the unified interface. Discrete tokenization integrates naturally with autoregressive modeling, enabling a single tokenized interface that unifies understanding and generation while remaining fully compatible with the LLM ecosystem~\citep{chen2025blip3o}.

\textbf{The bottleneck.} Compared to continuous features, discrete tokens typically have lower information density. For non-text modalities such as images, this creates a tight information bottleneck that can degrade multimodal understanding.

\textbf{Our approach.} We address this bottleneck with a product-quantization-based discretization scheme that substantially increases the information capacity of discrete vision tokens, yielding understanding on par with continuous-feature VLMs while remaining fully discrete and autoregressive. For the visual modality, we use a ViT-based encoder as the Tokenizer and a diffusion-based decoder as the Detokenizer, letting each component focus on its strength: the Tokenizer aligns visual signals into the shared semantic space, the LLM handles reasoning, and the Detokenizer renders high-quality images.

\section{Architecture}
\label{sec3_arch}


\begin{figure}[t!] 
  \centering
  \includegraphics[width=0.95\textwidth]{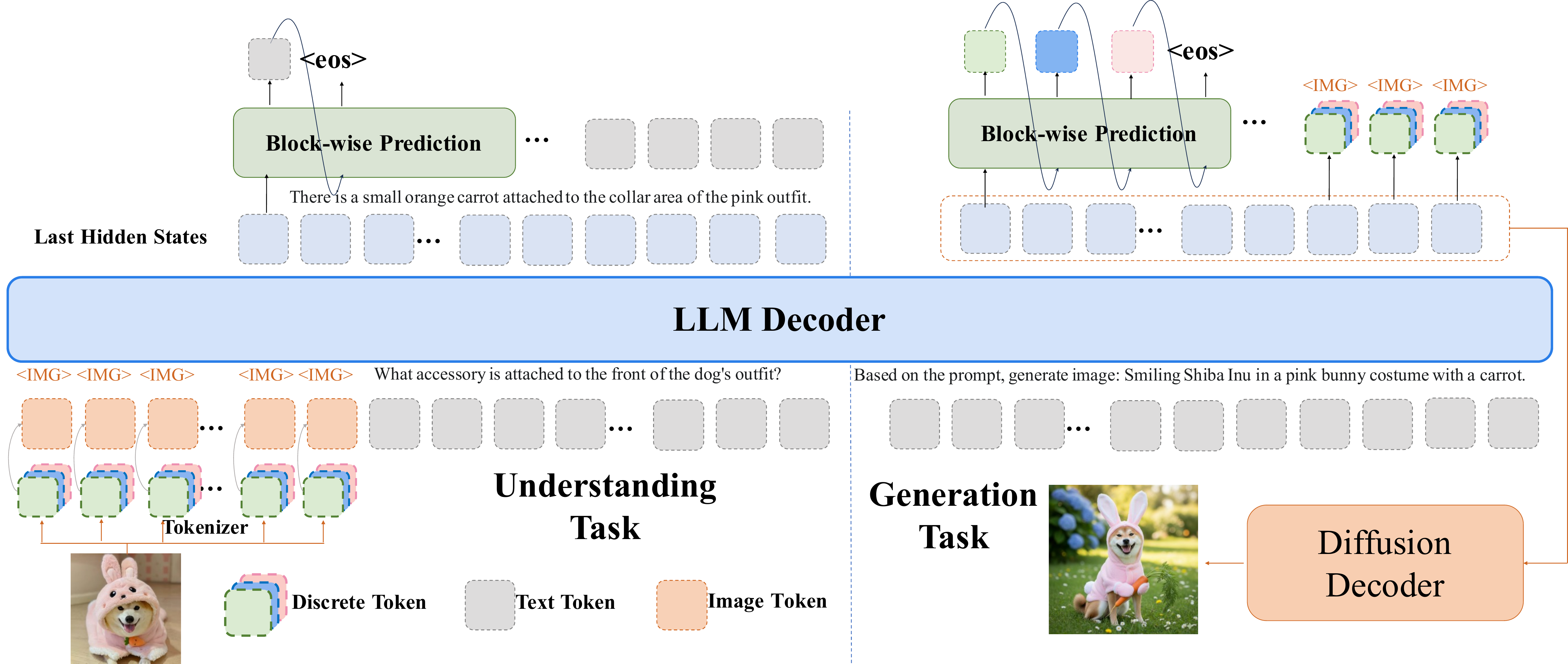}
    \caption{The auto-regressive inference workflow of our method, including the Kelix-Tok, and unified semantic understanding and image generation.}
    \label{mainmodel}
\end{figure}

Figure~\ref{mainmodel} shows a high-level training overview of our methods, which unify the semantic understanding\&image generation simultaneously.
In practice, our method consists by three major components:
\begin{itemize}
    \item \textbf{Kelix-Tokenizer}, which provides multiple parallel discrete tokens for each patch embedding.
    \item \textbf{Unified LLM Backbone}, to compress the high-level text and image semantic fusion.
    \item \textbf{Diffusion-based Image De-Tokenizer}, to render the high-resolution image according to last hidden states.
\end{itemize}
This section begins by introducing the discrete image tokenizer, detailing how to quantize the ViT-produced patch embeddings via a product quantization mechanism.
We then introduce a next-block-prediction training paradigm for the unified LLM backbone, which compresses the increased vision tokens from our multi-token tokenizer into manageable blocks, thereby preserving training and inference efficiency comparable to single-token methods.
Finally, we describe the diffusion-based de-tokenizer that renders high-quality images conditioned on the LLM-generated image tokens.

\begin{figure}[t!]
  \centering
  \includegraphics[width=0.95\textwidth]{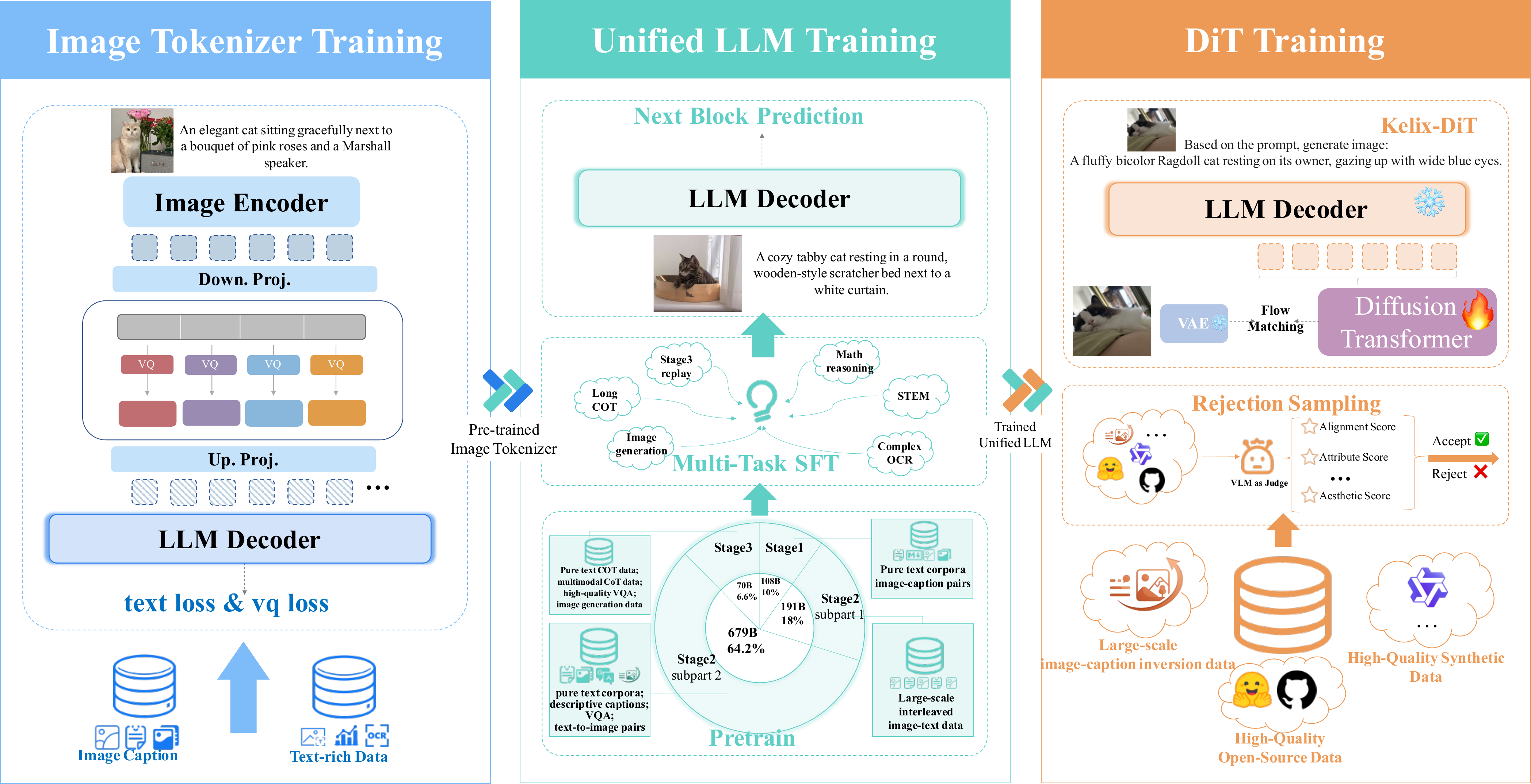}
    \caption{The auto-regressive training workflow of our method, including the Kelix Tokenizer, and Unified LLM model and Image DiT.}
    \label{train-pipeline}
\end{figure}

\begin{table}[t]
\footnotesize
	\centering
	\caption{A list of commonly used notation.}
	\setlength\tabcolsep{10pt}{
	\begin{tabular}{cl}
		\toprule
		\textbf{Notation} & \textbf{Description}\\
		\midrule
        \rowcolor{gray!10}
        \multicolumn{2}{l}{\textbf{Basic concept}} \\
		$I$ & An arbitrary image $I$. \\ \specialrule{0em}{2pt}{2pt}
        $P$ & The patch number of a Image $P$. \\ \specialrule{0em}{2pt}{2pt}
        $\mathbf{Z}/\mathbf{z}$ & The embedding matrix-of-all-patchs/vector-of-specific-patch. \\ \specialrule{0em}{2pt}{2pt}
        $\mathbf{W}$ & The learnable parameter matrix. \\ \specialrule{0em}{2pt}{2pt}
        \rowcolor{gray!10}
        \multicolumn{2}{l}{\textbf{VQ codebook concept}} \\
        $\mathbf{C}$ & The codebook embedding matrix. \\ \specialrule{0em}{2pt}{2pt}
        $d$ & The codebook dimension. \\ \specialrule{0em}{2pt}{2pt}
        $\mathbf{z}_q$ & The selected closest quantify embedding. \\ \specialrule{0em}{2pt}{2pt}
        \rowcolor{gray!10}
        \multicolumn{2}{l}{\textbf{Multiple VQ codebook concept}} \\
        $K$ & The codebook size, e.g., 8192. \\ \specialrule{0em}{2pt}{2pt}
        $N$ & The sub-codebook number of our Kelix, e.g., 8. \\ \specialrule{0em}{2pt}{2pt}
        $S$ & The sum of sub-codebook size of our Kelix, e.g., 8 $\times$ 8192 = 65536. \\ \specialrule{0em}{2pt}{2pt}
        \rowcolor{gray!10}
        \multicolumn{2}{l}{\textbf{Block prediction concept}} \\
        $t$ & The basic input token ID.  \\ \specialrule{0em}{2pt}{2pt}
        $B$ & The block input unit, text block is $\{t^1,\texttt{<eob>}\}$, visual block is $\{t^1,\dots,t^8,\texttt{<eob>}\}$ \\ \specialrule{0em}{2pt}{2pt}
        $\texttt{<eob>}$ & Additional special token end-of-block. \\ \specialrule{0em}{2pt}{2pt}
        $M$ & The block size, text block is 2, , visual block is 9. \\ \specialrule{0em}{2pt}{2pt}
        $\mathbf{e}$ & The input token embedding. \\ \specialrule{0em}{2pt}{2pt}
        $\mathbf{h}$ & The last hidden states. \\ \specialrule{0em}{2pt}{2pt}
		\midrule
	\bottomrule
	\end{tabular}
	\label{tab::symbol}
	}
\end{table}

\subsection{Discrete Image Tokenizer}
\label{sec:discrete_image_tokenizer}
Recent works on discrete image tokenizers, such as TAToK~\citep{kim2025democratizing}, BLIP3o-Next~\citep{chen2025blip3o}, Emu3~\citep{wang2024emu3}, and DualToken~\citep{song2025dualtoken}, X-omni~\cite{geng2025xomnireinforcementlearningmakes}, have demonstrated strong image generation capabilities.
However, these methods generally underperform VLMs that operate on continuous vision features in multi-modal understanding benchmarks (i.e., OCRbench, most recent VLMs achieve 85+, while the stongest unified model with discrete tokens only reach 70.). We attribute this gap to the inherent information bottleneck of single-token quantization. Consider, for example, a 1024-dimensional continuous embedding quantized into a single discrete token with a vocabulary size of 65{,}536: the resulting coding capacity is merely $\log_2 65536 = 16$ bits, whereas the original floating-point embedding occupies approximately 2\,KB. Although continuous embeddings inevitably contain some redundancy, their effective information capacity still far exceeds what 16 bits can represent.
To bridge this gap, we propose quantizing each continuous patch embedding into \emph{multiple} parallel discrete tokens, thereby significantly expanding the representational capacity. After quantization, we fuse these tokens back into a single composite representation before feeding it to the LLM decoder, which avoids the context length explosion that would otherwise result from the increased token count.
In the following, we describe the core design of Kelix-Tok in detail. An overview of the framework is presented on the left of Figure~\ref{train-pipeline}.

\textbf{Vector Quantization.}
Vector Quantization (VQ)~\citep{van2017neural} discretizes continuous representations into a finite set of learnable codes. Given an image $I$ with $P$ patches, a vision encoder produces a continuous feature matrix $\mathbf{Z}_{I} = [\mathbf{z}_{1}, \mathbf{z}_{2}, \cdots, \mathbf{z}_{P}] = \mathrm{ViT}(I) \in \mathbb{R}^{P \times d}$, where $d$ is the embedding dimension. Standard VQ maps each patch feature $\mathbf{z}$ to its nearest entry in a codebook $\mathbf{C} = \{\mathbf{c}_{1}, \mathbf{c}_{2}, \ldots, \mathbf{c}_{K}\}$:
\begin{equation}
    \mathbf{z}_{q} = \mathbf{c}_{\texttt{Index}}, \quad \texttt{Index} = \underset{\mathbf{c} \in \mathbf{C}}{\arg\min} \big\| \mathbf{z} - \mathbf{c} \big\|^{2}
\end{equation}
where $K$ is the codebook size. As discussed above, this single-token quantization creates a severe information bottleneck that the coding capacity of a single discrete index is far below the effective information carried by the original continuous embedding.

\textbf{Product Quantization for Visual Tokenization.}
To address this bottleneck, we adopt a product quantization (PQ)~\citep{jegou2011product} strategy for visual discretization: each continuous patch feature is decomposed into $N$ parallel subspaces via learned projection matrices, and quantized independently with its own sub-codebook. The combinatorial coding capacity thus grows exponentially with $N$, substantially narrowing the information gap with continuous representations.
To keep the context length unchanged for the LLM encoder, we apply \textbf{sum pooling
}to aggregate the $N$ quantized sub-vectors back into a single composite token on the encoder side, while the decoder retains the full $N$-way discrete codes for autoregressive prediction. This asymmetric design balances representational expressivity with computational efficiency.

\textbf{Codebook Initialization.}
Rather than random initialization, we leverage a pre-trained Keye-VL (NaViT) vision encoder to extract visual patch embedding from a large image corpus (more than 10,000,000 images), and apply K-means clustering to obtain $S = 65{,}536$ high-quality cluster centers. These centers are then uniformly partitioned into $N$ non-overlapping sub-codebooks, each containing $S / N$ entries, ensuring complementary semantic coverage across subspaces. During training, we adopt the SimVQ~\citep{zhu2025addressing} strategy: the cluster centers are frozen and only the per-subspace SimVQ projection matrices $\mathbf{W}_i$ are optimized, which maximizes codebook utilization and effectively mitigates codebook collapse.

\textbf{Architecture and Training Objective.}
Kelix-Tok comprises two core components: (1)~a NaViT-style vision encoder, initialized from Keye-VL and supporting native-resolution input, and (2)~the product quantization layer described above, which decomposes each continuous patch feature into $N$ parallel discrete codes.
To align the resulting discrete visual representations with the LLM text space, we attach a pre-trained lightweight language model (600M parameters) as an auxiliary decoder during training. The overall objective jointly optimizes a next-token prediction loss and a quantization loss:
\begin{equation}
\mathcal{L} = \mathcal{L}_{\text{NTP}} + \mathcal{L}_{\text{VQ}}
\end{equation}
where $\mathcal{L}_{\text{NTP}}$ is the standard cross-entropy loss that supervises the auxiliary decoder, enforcing semantic alignment between discrete visual tokens and the language space. The quantization loss $\mathcal{L}_{\text{VQ}}$ is defined as:
\begin{equation}
\mathcal{L}_{\text{VQ}} = \dfrac{1}{N}\sum_{i=1}^{N} \left( \underbrace{\big\| \text{sg}[\mathbf{z}^{\left(i\right)}] - \mathbf{z}_q^{\left(i\right)} \big\|_2^2}_{\text{Codebook Loss}} + \underbrace{\beta \big\| \mathbf{z}^{\left(i\right)} - \text{sg}[\mathbf{z}_q^{\left(i\right)}] \big\|_2^2 }_{\text{Commitment Loss}}\right)
\end{equation}
where $\mathbf{z}^{(i)}$ and $\mathbf{z}_q^{(i)}$ denote the continuous and quantized features from the $i$-th subspace, respectively, $\text{sg}[\cdot]$ is the stop-gradient operator, and $\beta$ controls the commitment weight. Intuitively, the codebook loss pulls each codebook entry toward the corresponding encoder output, while the commitment loss encourages the encoder to produce features that stay close to their quantized counterparts.

\textbf{Training Strategy.}
We adopt a staged unfreezing schedule: model parameters are gradually unlocked and trained on a diverse collection of understanding-oriented datasets spanning commonsense reasoning, inference, and text-intensive tasks. This curriculum strengthens the tokenizer's semantic comprehension while preserving the perceptual capabilities inherited from NaViT. Once training is complete, the auxiliary language model decoder is discarded; only the vision tokenizer is retained and integrated into the downstream larger unified LLM and image de-tokenizer pipeline.

\subsection{Unified Model Architecture}

As discussed in Section~\ref{sec:discrete_image_tokenizer}, our product quantization produces $N$ discrete codes per patch, substantially increasing the number of vision tokens compared to the original ViT output. To handle this without inflating the context length, we adopt a unified autoregressive Transformer architecture built around a \textbf{Next-Block Prediction (NBP)} paradigm.

\textbf{Next-Block Prediction (NBP).}
We organize multimodal sequences into a series of blocks $\{B_1, B_2, \dots\}$, where each block $B_i$ represents a coherent semantic unit terminated by an end-of-block token (\texttt{<eob>}). The block size is modality-dependent: 
\begin{itemize}[noitemsep, topsep=2pt]
    \item \myorange{\textbf{Text block ($\{t^1, \texttt{<eob>}\}$)}}: contains 1 text token plus \texttt{<eob>} ($|B|=2$), preserving token-level granularity consistent with standard text LLMs.
    \item \myorange{\textbf{Visual block ($\{t^1, \dots, t^N, \texttt{<eob>}\}$)}}: contains $N$ visual tokens plus \texttt{<eob>} ($|B|=N{+}1$), compressing the expanded discrete codes of each patch into a single prediction step while retaining local spatial structure.
\end{itemize}
At each autoregressive step $i$, the LLM processes block-level representations $\{B_1, \dots, B_i\}$ and predicts the entire next block $B_{i+1}$, effectively reducing the sequence length seen by the backbone by a factor of $N$ for visual data.

\textbf{Block-wise Input and Output Modules.}
To interface the LLM backbone with block-based sequences, we introduce two lightweight components: the \textit{Block Encoder} and the \textit{Block Decoder}. \myorange{For notational brevity, we denote each block as $B_i = \{t_{i}^1, t_{i}^2, \dots, t_{i}^M\}$, where $M = 2$ for text blocks and $M = N{+}1$ for visual blocks.}

\textit{Block Encoder (Embedding Aggregation):}
The \textit{Block Encoder} maps each block into a single fixed-dimensional representation for the LLM. Given the token embeddings $\{\mathbf{e}_{i}^1, \mathbf{e}_{i}^2, \dots, \mathbf{e}_{i}^M\}$ of block $B_i$, we compute the element-wise sum \myred{excluding the end-of-block embedding $\mathbf{e}_i^M$}:
\begin{equation}
\label{eq:ea_1}
    \mathbf{E}_i = \sum_{j=1}^{M-1} \mathbf{e}_{i}^j.
\end{equation}
This yields a modality-agnostic input to the LLM. The forward pass is then:
\begin{equation}
\label{eq:ea}
    \{\mathbf{h}_1,\dots,\mathbf{h}_{i}\} = \texttt{LLM}\left(\{\mathbf{E}_1,\dots,\mathbf{E}_{i}\}\right)
\end{equation}
where $\mathbf{h}_i$ denotes the last-layer hidden state at position $i$.

\textit{Block Decoder (Autoregressive Reconstruction):}
Given the hidden state $\mathbf{h}_i$, which encodes all preceding blocks $B_{\leq i}$, a lightweight Transformer decoder autoregressively reconstructs the next block:
\begin{equation}
    B_{i+1} = \texttt{BlockDecoder}\left(\mathbf{h}_i\right).
    \label{blockdecoder}
\end{equation}
The corresponding training loss factorizes as:
\begin{align*}
\mathcal{L}_{\text{nbp}}(B_{i+1}) &= -\log P\left(B_{i+1}\big|B_{\leq i}\right) \\
                  &= - \sum_{j=1}^{M} \log P\left(t^{j}_{i+1}\big|\mathbf{h}_{i},\, t^{<j}_{i+1}\right)
\end{align*}
where $\mathcal{L}_{\text{nbp}}$ serves as the training objective for the unified model.

\textbf{Implementation Details.}
We use Qwen3-8B~\citep{yang2025qwen3} as the LLM backbone. To accommodate the discrete visual codes from Kelix-Tok, we expand the vocabulary by $S = 65{,}536$ additional entries and augment the LM head accordingly. With the NBP paradigm, the effective context length for visual data is reduced by a factor of $N$, yielding training and inference efficiency comparable to single-token approaches.

\subsection{Diffusion-based Image De-Tokenizer}
The unified model described above produces two types of output: text tokens and image tokens. Text tokens can be directly mapped back to words via the standard text tokenizer. Image tokens, however, encode high-level semantic information that cannot be converted into pixels through a simple deterministic decoder (e.g., a VAE decoder~\citep{yang2024cogvideox}).
To render high-quality images from these semantic representations, we employ a latent diffusion model as the image de-tokenizer. Specifically, given a generation prompt, the Kelix LLM autoregressively produces a sequence of last hidden states $\{\mathbf{h}_*\}$ corresponding to the image blocks. We use these continuous hidden states as the semantic condition to guide a Diffusion Transformer (DiT). Following the flow-matching paradigm, we fine-tune a customized DiT (based on SANA-DiT~\citep{xie2024sana}) to denoise in the latent space of a pre-trained VAE, conditioned on $\{\mathbf{h}_*\}$. An overview is shown on the right of Figure~\ref{train-pipeline}.

\begin{table}[t]
    \caption{
        Comprehensive comparison on a curated set of multimodal benchmarks. The release dates denote the time of arXiv submission or official blog release.
    }
    \centering
    \small
    \renewcommand{\arraystretch}{1.2}
    \resizebox{\textwidth}{!}{
    \begin{tabular}{l c  ccc  ccc  ccc}
    \toprule
    \multirow{2}{*}{\textbf{Model}} & \multirow{2}{*}{\textbf{Date}} & \multicolumn{3}{c}{\textbf{General Benchmarks}} & \multicolumn{3}{c}{\textbf{Knowledge Benchmarks}} & \multicolumn{3}{c}{\textbf{Text-rich Benchmarks}} \\
    
    \cmidrule(lr){3-5} \cmidrule(lr){6-8} \cmidrule(lr){9-11}
    
    & & {SEED-Bench} & {RealWorldQA} & {MMBench$_\text{EN}$} & {AI2D$_\text{test}$} & {MMMU} & {MathVista} & {ChartQA} & {TextVQA} & {OCRBench} \\
    \midrule
    
    \rowcolor{gray!10}
    \multicolumn{11}{l}{\textbf{Understanding-only Models}} \\
    LLaVA-OV (7B)         & 2024.08   & 76.7 & 69.9 & -- & 81.4 & 48.8 & -- & 80.9 & -- & 62.2 \\
    BLIP-3 (4B)           & 2024.08   & 72.2 & 60.5 & -- & --   & 41.1 & 39.6 & -- & 71.0 & -- \\
    InternVL3.5 (4B)      & 2025.08   & --   & 66.3 & -- & 82.6 & 66.6 & 77.1 & 86.0 & 77.9 & 82.2 \\
    Qwen2.5VL (3B)        & 2025.02   & --   & 65.4 & 76.4 & 81.6 & 53.1 & 62.3 & 84.0 & 79.7 & 79.7 \\
    Qwen2.5VL (7B)        & 2025.02   & --   & 68.5 & 82.6 & 83.9 & 58.6 & 68.2 & 87.3 & 84.9 & 86.4 \\
    \midrule

    \rowcolor{gray!10}
    \multicolumn{11}{l}{\textbf{Unified Multimodal LLMs}} \\
    
    \multicolumn{11}{c}{\textbf{\textit{Continuous Tokenization}}} \\
    Bagel (14B)           & 2025.05   & 78.5 & 72.8 & 85.0 & 89.2 & 55.3 & 73.1 & 78.5 & 80.0 & 78.3 \\
    Ming-Univision (16B)  & 2025.10   & --   & --   & --   & 82.8   & 40.3   & 66.6   & --   & --   & 72.4   \\
    Blip-3o (4B)          & 2025.05   & 73.8 & 60.4 & 78.6 & --   & 46.6 & --   & --   & 78.0 & --   \\
    TUNA (7B)             & 2025.12   & 74.7 & 66.1 & --   & 79.3 & 49.8 & --   & 85.8 & --   & 74.3 \\

    \multicolumn{11}{c}{\textbf{\textit{Hybrid Tokenization}}} \\
    Manzano (30B)         & 2025.09   & 76.0 & 70.1 & 83.4 & 86.0 & 57.8 & 73.3 & 89.0 & 84.3 & 86.3 \\
    VQ-RAE (7B)  & 2025.11   & 77.0 & -- & 85.1 & 84.8 & 61.6 & -- & -- & 80.6 & -- \\
    \rowcolor{blue!10}
    \multicolumn{11}{c}{\textbf{\textit{Discrete Tokenization}}} \\
    \rowcolor{blue!10}
    Janus-Pro (1.5B)      & 2025.01   & 68.3 & 52.6 & 75.5 & 64.5 & 36.3 & 36.5 & 23.4 & 41.9 & 48.7 \\
    \rowcolor{blue!10}
    Janus-Pro (7B)        & 2025.01   & 72.1 & 58.0 & \underline{79.2} & 71.3 & 41.0 & 42.5 & 25.8 & 45.6 & 59.0 \\
    \rowcolor{blue!10}
    Emu3 (8B)             & 2024.09   & 68.2 & 57.4 & 58.5 & 70.0 & 31.6 & 47.6 & --   & 64.7 & 68.7 \\
    \rowcolor{blue!10}
    X-Omni (7B)           & 2025.07   & \underline{74.3} & \underline{62.6} & 74.8 & 76.8 & 47.2 & \underline{54.1} & \underline{81.5} & \underline{77.4} & \underline{70.4} \\
    \rowcolor{blue!10}
    Show-O2 (7B)          & 2025.06   & 69.8   & --   & --   & \underline{78.6}   & \underline{48.9}   & --   & 66.9   & 71.5   & --   \\
    \rowcolor{blue!10}
    Unitok (7B)           & 2025.02   & --   & --   & --   & --   & --   & --   & --   & 51.6   & --   \\
    \rowcolor{blue!10}
    TokenFlow-XL (14B)       & 2024.12   & 72.6   & 56.6   & --   & 75.8   & 43.2   & --   & --   & 62.3   & --   \\
    \rowcolor{blue!10}
    Nextflow (7B)         & 2026.01   & --   & --   & --   & --   & 37.1   & --   & 57.7   & 58.9   & 55.1   \\
    \rowcolor{blue!10}
    Kelix (8B)          &           & \textbf{76.0} & \textbf{72.1} & \textbf{80.2} & \textbf{82.4} & \textbf{54.1} & \textbf{76.5} & \textbf{83.0} & \textbf{81.4} & \textbf{86.7} \\
    \midrule
    
    \rowcolor{gray!10}
    \multicolumn{11}{l}{\textbf{Proprietary Models}} \\
    GPT-4o                & 2024.05   & 77.1 & 75.4 & --   & 84.6 & 69.2 & 61.3 & 85.7 & --   & 73.6 \\
    Gemini-2.5-Pro        & 2024.12   & --   & 78.0 & 86.3 & 89.5 & 81.7 & 82.7 & 83.3 & 76.8 & 86.2 \\
    
    \bottomrule
    \end{tabular}
    }
\label{tab:full_comprehensive_evaluation_clean}
For a fair comparison, baselines highlighted with \colorbox{blue!10}{blue shading} adopt the same discrete tokenization paradigm and comparable model scale as Kelix. \textbf{Best} and \underline{runner-up} results among these comparable models are marked in \textbf{bold} and \underline{underlined}, respectively.
\end{table}

\begin{table}[t]
    \caption{
        Comprehensive comparison of image generation performance on GenEval and WISE benchmarks. The release dates denote the time of arXiv submission or official release. \textbf{For Kelix, we report performance using the default resolution of $1024\times1024$.}
    }
    \centering
    \small
    \renewcommand{\arraystretch}{1.2}
    \resizebox{\textwidth}{!}{
    \begin{tabular}{l c  ccccccc  ccccccc}
    \toprule
    \multirow{2}{*}{\textbf{Model}} & \multirow{2}{*}{\textbf{Date}} & \multicolumn{7}{c}{\textbf{GenEval Benchmark}} & \multicolumn{7}{c}{\textbf{WISE Benchmark}} \\
    \cmidrule(lr){3-9} \cmidrule(lr){10-16}
    & & Single & Two & Count & Color & Pos. & C-Attr & \textbf{Overall} & Cult. & Time & Space & Bio. & Phys. & Chem. & \textbf{Overall} \\
    \midrule
    
    \rowcolor{gray!10}
    \multicolumn{16}{l}{\textbf{Dedicated T2I Models}} \\
    SDXL (3.5B)           & 2023.07   & 98.0 & 74.0 & 39.0 & 85.0 & 15.0 & 23.0 & 55.0 & 43.0 & 48.0 & 47.0 & 44.0 & 45.0 & 27.0 & 43.0 \\
    DALL-E 3              & 2023.10   & 96.0 & 87.0 & 47.0 & 83.0 & 43.0 & 45.0 & 67.0 & --   & --   & --   & --   & --   & --   & --   \\
    SD3 Medium (2B)       & 2024.06   & 99.0 & 94.0 & 72.0 & 89.0 & 33.0 & 60.0 & 74.0 & 43.0 & 50.0 & 52.0 & 41.0 & 53.0 & 33.0 & 45.0 \\
    FLUX.1-dev (12B)      & 2024.08   & 98.0 & 93.0 & 75.0 & 93.0 & 68.0 & 65.0 & 82.0 & 48.0 & 58.0 & 62.0 & 42.0 & 51.0 & 35.0 & 50.0 \\
    \midrule

    \rowcolor{gray!10}
    \multicolumn{16}{l}{\textbf{LLM \& Diffusion Conjunction}} \\
    MetaQuery-XL (7B)     & 2025.05   & --   & --   & --   & --   & --   & --   & 80.0 & 56.0 & 55.0 & 62.0 & 49.0 & 63.0 & 41.0 & 55.0 \\
    OmniGen-2 (7B)        & 2025.06   & 100.0 & 95.0 & 64.0 & 88.0 & 55.0 & 76.0 & 80.0 & --   & --   & --   & --   & --   & --   & --   \\
    Qwen-Image (27B)      & 2025.08   & 99.0 & 92.0 & 89.0 & 88.0 & 76.0 & 77.0 & 87.0 & -- & -- & -- & -- & -- & -- & -- \\
    \midrule

    \rowcolor{gray!10}
    \multicolumn{16}{l}{\textbf{Unified Multimodal LLMs}} \\
    
    \multicolumn{16}{c}{\textbf{\textit{Continuous Tokenization}}} \\
    Bagel (14B)           & 2025.05   & 99.0 & 94.0 & 81.0 & 88.0 & 64.0 & 63.0 & 82.0 & 44.0 & 55.0 & 68.0 & 44.0 & 60.0 & 39.0 & 52.0 \\
    Ming-Univision (16B)  & 2025.10   & 100.0   & 93.0   & 59.0   & 93.0   & 92.0   & 70.0   & 85.0 & --   & --   & --   & --   & --   & --   & --   \\
    Blip-3o (4B)          & 2025.05   & --   & --   & --   & --   & --   & --   & 81.0 & --   & --   & --   & --   & --   & --   & 50.0 \\
    TUNA (7B)             & 2025.12   & 100   & 97   & 81   & 91   & 88   & 83   & 90.0 & --   & --   & --   & --   & --   & --   & --   \\

    \multicolumn{16}{c}{\textbf{\textit{Hybrid Tokenization}}} \\
    Manzano (30B)         & 2025.09   & 100.0 & 91.0 & 83.0 & 87.0 & 84.0 & 65.0 & 85.0 & 58.0 & 50.0 & 65.0 & 50.0 & 55.0 & 32.0 & 54.0 \\
    VQ-RAE (7B)  & 2025.11   & 96.0 & 82.0 & 64.0 & 80.0 & 73.0 & 58.0 & 76.0 & -- & -- & -- & -- & -- & -- & -- \\

    \rowcolor{blue!10}
    \multicolumn{16}{c}{\textbf{\textit{Discrete Tokenization}}} \\
    \rowcolor{blue!10}
    Janus-Pro (7B)        & 2025.01   & \underline{99.0} & 89.0 & 59.0 & 90.0 & \underline{79.0} & 66.0 & 80.0 & 30.0 & 37.0 & 49.0 & 36.0 & 42.0 & 26.0 & 35.0 \\
    \rowcolor{blue!10}
    Emu3 (8B)             & 2024.09   & \underline{99.0}   & 81.0   & 42.0   & 80.0   & 49.0   & 45.0   & 66.0 & --   & --   & --   & --   & --   & --   & 39.0 \\
    \rowcolor{blue!10}
    X-Omni (7B)           & 2025.07   & 98.0 & \textbf{95.0} & \underline{75.0} & \underline{91.0} & 71.0 & 68.0 & 83.0 & --   & --   & --   & --   & --   & --   & --   \\
    \rowcolor{blue!10}
    Show-O2 (7B)          & 2025.06   & \textbf{100.0}   & 87.0   & 58.0   & \textbf{92.0}   & 52.0   & 62.0 & 76.0 & --   & --   & --   & --   & --   & --   & --   \\
    \rowcolor{blue!10}
    Unitok (7B)           & 2025.02   & \underline{99.0}   & 71.0   & 36.0   & 79.0   & 26.0   & 45.0   & 59.0 & --   & --   & --   & --   & --   & --   & --   \\
    \rowcolor{blue!10}
    TokenFlow (13B)       & 2024.12   & 97.0   & 66.0   & 40.0   & 84.0   & 17.0   & 26.0   & 55.0 & --   & --   & --   & --   & --   & --   & --   \\
    \rowcolor{blue!10}
    Nextflow (7B)         & 2026.01   & 98.0   & 92.0   & 73.0   & 90.0   & 77.0   & \underline{69.0}   & 83.0 & \textbf{62.0}   & \underline{60.0}   & \textbf{70.0}   & \textbf{54.0}   & \underline{58.0}   & \textbf{38.0}   & \textbf{59.0} \\
    \rowcolor{blue!10}
    \rowcolor{blue!10}
    Kelix (8B) & & \textbf{100.0} & \underline{93.0} & \textbf{81.0} & 87.0 & \textbf{85.0} & \textbf{79.0} & \textbf{87.6} & \underline{58.0} & \textbf{62.0} & \underline{68.0} & \underline{47.0} & \textbf{61.0} & \underline{35.0} & \underline{57.0} \\
    \midrule
    
    \rowcolor{gray!10}
    \multicolumn{16}{l}{\textbf{Proprietary Models}} \\
    GPT-4o                & 2024.05   & 99.0 & 92.0 & 85.0 & 92.0 & 75.0 & 61.0 & 84.0 & 81.0 & 71.0 & 89.0 & 83.0 & 79.0 & 74.0 & 80.0 \\
    
    \bottomrule
    \end{tabular}
    }
    \label{tab:generation_results}
\end{table}

\begin{table}[t!]
\footnotesize
    \caption{
        Comparison of image generation performance on the DPG-Bench benchmark. All scores are reported on a 100-point scale. The release dates denote the time of arXiv submission or official release.
    }
    \centering
    \small
    \renewcommand{\arraystretch}{1.2}
    
    \begin{tabular}{l c  ccccc  c}
    \toprule
    \textbf{Method} & \textbf{Date} & \textbf{Global} & \textbf{Entity} & \textbf{Attribute} & \textbf{Relation} & \textbf{Other} & \textbf{Overall$\uparrow$} \\
    \midrule
    
    \rowcolor{gray!10}
    \multicolumn{8}{l}{\textbf{Dedicated T2I Models}} \\
    SDXL (2.6B)           & 2023.07 & 83.3 & 82.4 & 80.9 & 86.8 & 80.4 & 74.7 \\
    DALL-E 3              & 2023.10 & 91.0 & 89.6 & 88.4 & 90.6 & 89.8 & 83.5 \\
    SD3 Medium (2B)       & 2024.06 & 87.9 & 91.0 & 88.8 & 80.7 & 88.7 & 84.1 \\
    FLUX.1-dev (12B)      & 2024.08 & --   & --   & --   & --   & --   & --   \\
    \midrule

    \rowcolor{gray!10}
    \multicolumn{8}{l}{\textbf{LLM \& Diffusion Conjunction}} \\
    MetaQuery-XL (7B)     & 2025.04 & --   & --   & --   & --   & --   & 82.05  \\
    OmniGen-2 (7B)        & 2025.06 & 88.81   & 88.83   & 90.18   & 89.37   & 90.27   & 83.57   \\
    Qwen-Image (27B)      & 2025.08 & 91.32   & 91.56   & 92.02   & 94.31   & 92.73   & 88.32   \\
    \midrule

    \rowcolor{gray!10}
    \multicolumn{8}{l}{\textbf{Unified Multimodal LLMs}} \\
    
    \multicolumn{8}{c}{\textbf{\textit{Continuous Tokenization}}} \\
    Bagel (14B)           & 2025.05 & --   & --   & --   & --   & --   & --   \\
    Ming-Univision (16B)  & 2025.10 & --   & --   & --   & --   & --   & 82.1 \\
    Blip-3o (4B)          & 2025.05 & --   & --   & --   & --   & --   & 79.4 \\
    TUNA (7B)             & 2025.12 & 90.42   & 91.58   & 90.94   & 91.87   & 90.73   & 86.8 \\

    \multicolumn{8}{c}{\textbf{\textit{Hybrid Tokenization}}} \\
    Manzano (30B)         & 2025.09 & --   & --   & --   & --   & --   & --   \\
    VQ-RAE (7B)         & 2025.11 & 89.8   & 93.1   & 89.9   & 90.3   & 91.2   & 86.6   \\

    \rowcolor{blue!10}
    \multicolumn{8}{c}{\textbf{\textit{Discrete Tokenization}}} \\
    \rowcolor{blue!10}
    Janus-Pro (7B)        & 2025.01 & 86.9   & 88.9   & 89.4   & 89.3   & 89.5   & 84.2 \\
    \rowcolor{blue!10}
    Emu3 (8B)             & 2024.09 & 85.2 & 86.7 & 86.8 & 90.2 & 83.2 & 80.6 \\
    \rowcolor{blue!10}
    X-Omni (7B)           & 2025.07 & 84.8   & \textbf{92.6}   & \textbf{90.6}   & \textbf{94.8}   & 84.2   & \textbf{87.7} \\
    \rowcolor{blue!10}
    Show-O2 (7B)          & 2025.06 &  \underline{89.0}  & \underline{91.8}   & 90.0   & 91.9   & \textbf{91.6}   & \underline{86.1} \\
    \rowcolor{blue!10}
    Unitok (7B)           & 2025.02 & --   & --   & --   & --   & --   & -- \\
    \rowcolor{blue!10}
    TokenFlow (13B)       & 2024.12 & 78.7 & 79.2 & 81.3 & 85.2 & 71.2 & 73.4 \\
    \rowcolor{blue!10}
    Nextflow (7B)         & 2026.01 & \textbf{92.4}   & 90.1   & \underline{90.5}   & 92.7   & \underline{91.1}   & 86.0 \\
    \rowcolor{blue!10}
    \rowcolor{blue!10}
    Kelix (8B)          &       & 84.5 & 91.5 & 89.2 & \underline{93.0} & 84.4 & 85.5 \\
    
    

    \bottomrule
    \end{tabular}
    \label{tab:dpg_bench_results}
\end{table}

\section{Training}

In this section, we describe the dataset construction, stage-wise training configurations, evaluation benchmark results, and detailed ablation studies.

\subsection{Datasets}
We curate a large-scale, diverse, and high-quality corpus exceeding 1T tokens to train the full Kelix pipeline, spanning the discrete image tokenizer, the unified LLM, and the diffusion-based image de-tokenizer. Our training data covers the following categories:

\begin{itemize}
    \item \textbf{Image Caption}: Large-scale image-text pairs that establish foundational world knowledge by mapping visual features to linguistic concepts. Sources include Coyo~\citep{kakaobrain2022coyo-700m}, DataComp~\citep{gadre2023datacomp}, LAION~\citep{schuhmann2022laion}, Infinity-MM Stage1~\citep{gu2024infinity}, CC12M~\citep{changpinyo2021conceptual}, PD12M~\citep{meyer2024public}, ConceptualCaptions~\citep{sharma2018conceptual}, DenseFusion~\citep{li2024DenseFusion}, among others.
    \item \textbf{Optical Character Recognition}: OCR data spanning handwritten text, charts, synthetic documents, and multi-image layouts, sourced from OneVision~\citep{zheng2025onevision}, CodeOCR~\citep{codeocr_leetcode_2025}, SynthDoG~\citep{kim2022ocr}, InstructOCR~\citep{duan2025instructocr}, CCPD~\citep{xu2018towards}, and others. To address the scarcity of Chinese OCR data, we synthesize in-house datasets with multi-turn QA tasks covering diverse character sets, non-conventional reading orders, and specialized layouts (\textit{e.g.}, ancient Chinese poems), thereby strengthening the model's ability to parse spatial arrangement, font variation, and instruction-following.To enhance the model’s OCR grounding and performance on text-dense tasks, we apply rejection sampling on natural images to select samples that are (i) text-heavy, (ii) contain small and hard-to-read text, and (iii) exhibit sparsely and discretely distributed text regions. Using Qwen3-VL and Gemini, we then construct multi-dimensional OCR-focused QA pairs. In addition, since document OCR capability is critical, we incorporate training data spanning multiple programming languages as well as HTML, PDF, and Markdown formats.

\begin{figure}[t!]
  \centering
  \includegraphics[width=0.9\textwidth]{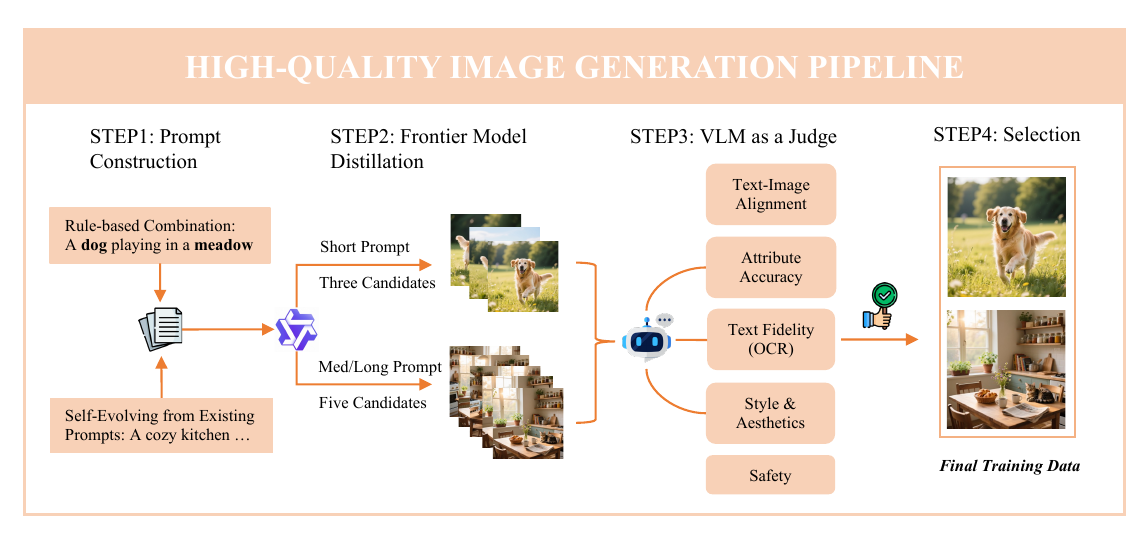}
    \caption{Overview of the high-quality text-to-image training data synthesis pipeline, comprising prompt construction, frontier model distillation, VLM-based multi-dimensional quality verification, and final selection.}
    \label{fig:generation_data_pipeline}
\end{figure}
    
    \item \textbf{Visual Question Answering}: VQA data from diverse domains to strengthen context-aware visual reasoning, including ShareGPT-4o~\citep{cui2025comprehensive}, Docmatix~\citep{laurencon2024building}, M4-instruct~\citep{li2024llavanext-interleave}, ShareGPT4V~\citep{chen2024sharegpt4v}, DVQA~\citep{kafle2018dvqa}, Mavis-Function~\citep{zhang2024mavis}, Honey-Bee~\citep{bansal2025honeybee}, FineVision~\citep{wiedmann2025finevision}, etc.
    \item \textbf{STEM \& Interleaved}: Interleaved image-text data for long-context multimodal modeling, drawn from Wikipedia~\citep{wikidump}, ArxivPapers, MM-Interleaved~\citep{tian2024mm}, and Pdf2Doc.
    \item \textbf{Pure Text}: A large-scale text corpus spanning math, code, chain-of-thought reasoning, and common knowledge, used to preserve the backbone's pretrained capabilities. Sources include Nemotron-CC~\citep{nvidia_nemotron_nano_v3_2025}, Nemotron-CC-Math~\citep{karimi2025nemotroncc,nvidia2025nvidianemotronnano2}, GeneralThought-430K~\citep{RJT1990_2025GeneralThoughtArchive}, NuminaMath-QwQ-CoT-5M~\citep{primeintellect2024numinamath}, and OpenCodeReasoning~\citep{ahmad2025opencodereasoning}.
    \item \textbf{Text-to-Image}: For image generation, we curate text-to-image paired data from CommonCatalog~\citep{gokaslan2024commoncanvas}, midjourney-niji-1m-llavanext~\citep{midjourney-niji-1m-llavanext}, BLIP3o-Pretrain-Short-Caption and BLIP3o-Pretrain-Long-Caption~\citep{chen2025blip3}, along with inverted caption datasets. Beyond public sources, we synthesize in-house data targeting fine-grained control over object attributes, spatial relations, counting, and text rendering. As illustrated in Figure~\ref{fig:generation_data_pipeline}, we first construct seed prompts at multiple complexity tiers (rule-based short prompts and LLM-generated medium/long prompts), then generate candidate images via frontier open-source models (Qwen-Image) with best-of-N sampling, and finally apply multi-dimensional VLM-based quality verification (text--image alignment, attribute accuracy, text fidelity, style, and safety) using Qwen2.5-VL-72B. Rejected images with acceptable visual quality are recaptioned to recover valid training pairs.
\end{itemize}

\subsection{Kelix-Tokenizer Training Recipes}

\begin{figure}[t!]
  \centering
  \includegraphics[width=0.75\textwidth]{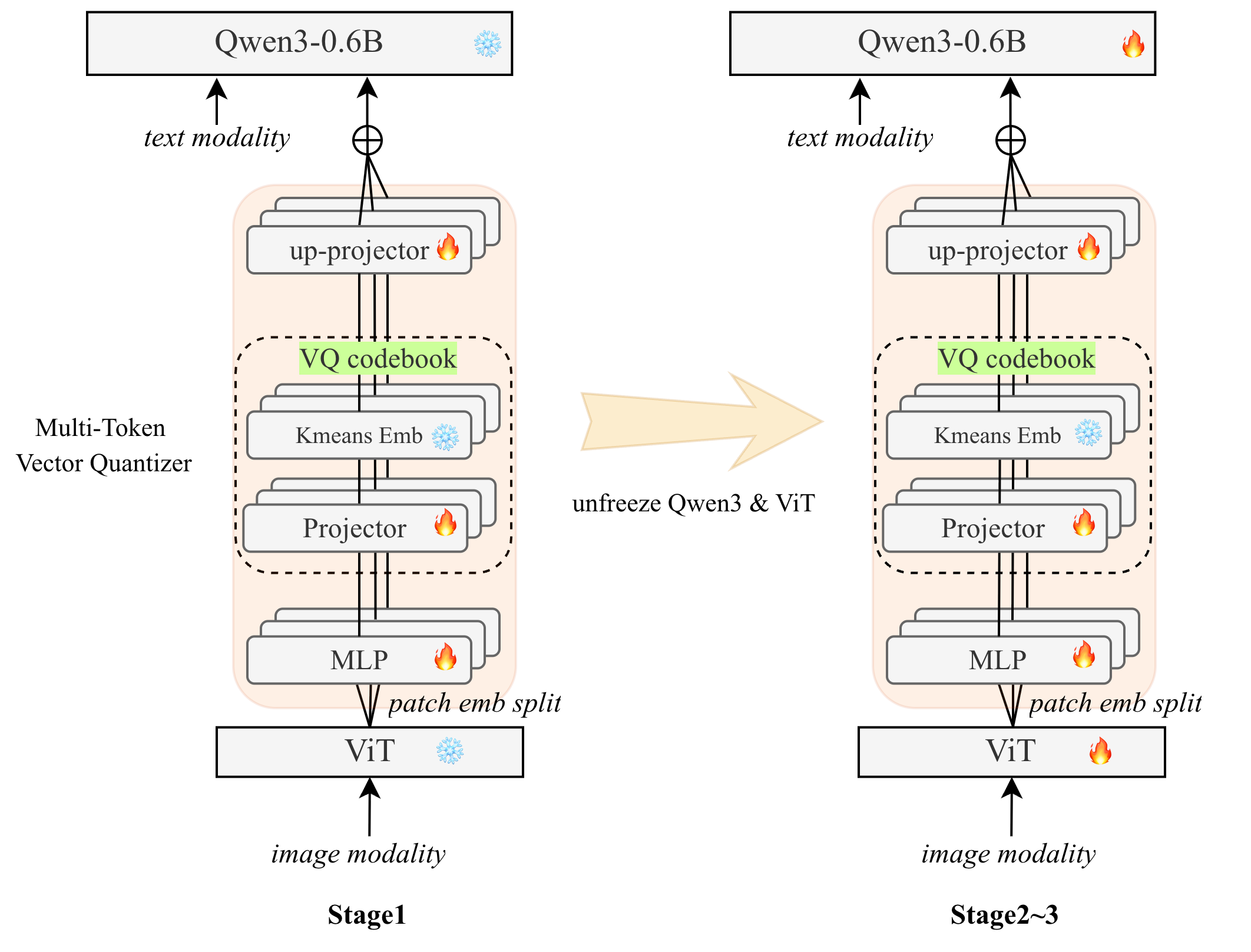}
    \caption{Semantic tokenizer training recipe. In Stage 1, we freeze the ViT and LLM backbones. To ensure that the commitment loss can be optimized, we insert the MLP and SimVQ up-projector before VQ codebook. In Stage2 and Stage3, we enable the gradients of ViT and Qwen3.}
    \label{fig:tok-recipe}
\end{figure}

Figure \ref{fig:tok-recipe} shows the training recipes of Kelix-Tok.
The backbone of Kelix-Tok is initialized from Keye-VL 1.5~\citep{team2025kwai}'s pre-trained naitive resolution ViT (NaViT). To enhance the vision-language semantic alignment, we connect the ViT to a small LLM, i.e., Qwen3-0.6B ~\citep{yang2025qwen3}.
\subsubsection{Codebook Initialization}
For VQ codebook initialization, we first leverage the pre-trained NaViT to extract patch embeddings from a large-scale image corpus, then perform K-means clustering on these embeddings to obtain high-quality cluster centers as the initial representations of the VQ codebook. These cluster centers are further randomly partitioned into N equal size non-overlapping sub-codebooks to ensure complementary semantic coverage across different subspaces. To achieve stable and fast convergence of codebook training, we freeze the centroid embeddings derived from K-means clustering and adopt the SimVQ~\citep{zhu2025addressing} training strategy to optimize the codebook. This strategy only updates the per-subspace projection matrices, which maximizes codebook utilization and effectively mitigates the codebook collapse problem.

\subsubsection{Training Pipline}
Kelix-Tok is optimized through a meticulously designed three-stage training pipeline, with each stage tailored to distinct training objectives to progressively boost the tokenizer's cross-modal semantic encoding capability, as detailed below:

\paragraph{Stage 1: Codebook Alignment} %
This stage solely trains the multi-discrete adapters with image caption corpora, while the core NaViT encoder and Qwen3-0.6B LLM backbone remain frozen. This stage enables the VQ codebook to establish an initial visual-semantic mapping, laying a fundamental foundation for subsequent cross-modal alignment between visual and linguistic representations.

\paragraph{Stage 2: Multi-Task Pre-training} %
We unlock all model parameters for end-to-end training on a diverse collection of understanding-oriented datasets, covering general knowledge, logical reasoning, and text-rich tasks. By exposing the tokenizer to a broad data distribution, this stage fully explores the storage capacity of the codebook and expands its knowledge reservoir. Meanwhile, it adapts the tokenizer to the autoregressive generation paradigm of LLMs, achieving deep alignment between discrete visual representations and the LLM's semantic space.

\paragraph{Stage 3: Annealing Training} %
As the final fine-tuning stage, we optimize the tokenizer on high-quality instruction-tuning datasets with an annealing learning rate. The core objective is to capture the inherent underlying rules and latent semantic information from the data, further enhancing the expressiveness of discrete visual representations and the robustness of cross-modal reasoning.

Upon the completion of the entire training pipeline, the auxiliary Qwen3-0.6B LLM decoder is discarded. The resulting Kelix-Tok, an integration of the NaViT encoder and a set of VQ codebooks, serves as a dedicated image tokenizer that encodes raw images into structured discrete visual representations. This design allows the knowledge of the image domain to be seamlessly transferred to various LLM decoders, realizing efficient cross-modal knowledge sharing. %

\subsection{Kelix-LLM Training Recipes}
\setlength{\leftmargin}{\parindent}
\setlength{\itemindent}{0pt}
\setlength{\labelwidth}{0pt}
\label{sec:training_strategy}

\begin{figure}[t!]
  \centering
  \includegraphics[width=0.75\textwidth]{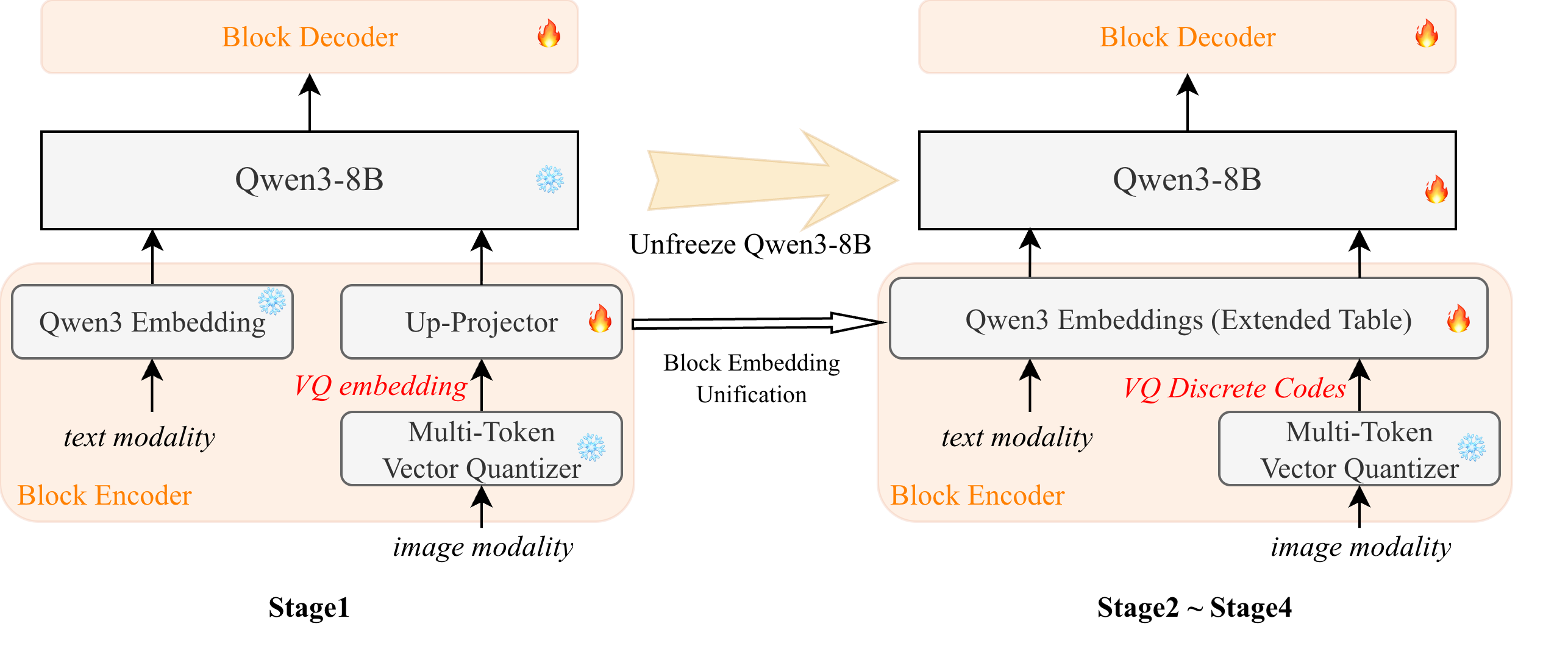}
    \caption{The LLM training recipe of Kelix, we first pre-train the \textit{Block Decoder} and the image token embedding for a better warm start, and then unifying the image token and text token to a same embedding table for full-parameter tuning.}
    \label{fig:ar-recipe}
\end{figure}

The proposed unified model follows a four-stage progressive training strategy to bridge textual and visual modalities while preserving the core reasoning capabilities of the LLM. 
This LLM training protocol progressively optimizes different model components, with data scale and composition  tailored to the objectives of each stage. 
Table \ref{tab:ar_training_data} summarizes the number of samples and token volume for each training stage, illustrating the progressive increase in data complexity and the shift toward high-quality, task-specific data in later stages.

\subsubsection{Visual Processor}
To enhance spatial position awareness under Qwen3’s 1D Rotary Position Embedding (RoPE) framework, we introduce a dedicated visual processor based on Keye-VL1.5~\citep{yang2025kwai}, uniformly applied to both image understanding and generation.

With the patch tokens given by Kelix-Tok, each token row is prepended with special tokens and embedded with ``row, column'' spatial coordinates (\textit{e.g.}, <|mm\_pos\_start|>3,4<|mm\_pos\_end|>). These markers extend 1D RoPE with structured 2D spatial information, allowing accurate modeling of inter-patch positional relationships while preserving the original Transformer architecture.

\subsubsection{Stage 1: Cross-Modal Alignment}
This stage serves as a ``warm-up'' to establish a stable mapping between discrete visual representations and the LLM’s linguistic latent space, avoiding catastrophic forgetting of pre-trained linguistic priors. All images are processed at native resolution, as illustrated in Stage 1 on the left of Figure \ref{fig:ar-recipe}.

In \textit{Block Encoder}, the textual modality is mapped to textual embeddings via \textit{Qwen3-8B Embedding}; the visual modality is encoded into VQ embeddings by Kelix-Tok. To match the LLM backbone’s dimensionality, the VQ embedding is transformed via a linear \textit{up-projector}; Decoder layers and embedding layer of Qwen3-8B are fully frozen; only the \textit{Block Decoder} and LM head are optimized, with training data comprising large-scale pure text corpora and image-caption pairs. 

This stage lays the foundation for cross-modal semantic consistency by completing the initial alignment between visual and textual representations.

\subsubsection{Stage 2: Unified Multimodal Pre-Training}
Building on Stage 1, this stage integrates embedding space unification, cross-modal generalization scaling, and multi-task capability development. It transforms the LLM into a native multimodal engine with integrated understanding, reasoning, and generation capabilities through a two-substage progressive training process.

\paragraph{Embedding Unification}
Following Stage 1, the VQ embedding table is projected sequentially by two pre-trained \textit{up-projectors} to generate a unified image embedding table: the first is the pre-trained \textit{up-projector} from the Kelix-Tok, and the second is the \textit{up-projector} fine-tuned in LLM Stage 1. This image embedding table is then concatenated with Qwen3’s text embedding table to form a unified \textit{Block Encoder}, which replaces the separate Qwen3 Embedding, VQ Embedding, and \textit{up-projector} components used in Stage 1. 

\paragraph{Interleaved Scaling (Stage 2.1)}
This substage focuses on achieving seamless integration of textual and visual modalities by unifying the embedding space, while expanding data diversity to enhance cross-modal generalization. All model parameters are unfrozen to enable end-to-end autoregressive training.
Training data utilizes large-scale interleaved image-text data and text corpora to balance cross-modal scaling and linguistic capability preservation.

This substage lays the structural and data foundation for complex multi-task learning by unifying the embedding space and expanding cross-modal data coverage.

\paragraph{Multi-Task Pre-Training (Stage 2.2) }
Building on the previous substage, this substage enriches the model’s multimodal capabilities through diverse task data, shaping basic understanding, reasoning, and generation competencies. Training relies on a large-scale, diverse corpus covering pure text corpora, descriptive captions, simple visual-language QA, and text-to-image pairs.
For image generation tasks, input images are constrained to an aspect ratio range of 0.75 to 1.33 and then resized to a fixed resolution of 504×504 pixels. We also enable the completion-only loss mask that backpropagates negative log-likelihood loss solely through the assistant’s response tokens. 

This substage equips the model with comprehensive multimodal capabilities by integrating diverse task data, balancing understanding and generation performance for subsequent refinement.

\subsubsection{Stage 3: Annealing Training}
This stage focuses on enhancing instruction following, logical reasoning, and generative quality, with learning rate annealing initiated to ensure stable convergence. Curated high-quality training data includes CoT data, high-quality filtered multimodal QA, filtered text corpora, and general image generation data, with rejection sampling applied to filter low-quality data, prevent hallucinations, and enhance fine-grained abilities.

This stage refines the model’s reasoning rigor and generative stability, laying the foundation for task-specific optimization.

\subsubsection{Stage 4: Supervised Fine-Tuning (SFT)}
Building directly on Stage 3, this stage retains the annealed learning rate and focuses on optimizing task-specific performance while preserving unified multimodal capabilities. Core training data includes replayed high-quality Stage 3 data, task-specific CoT data, targeted image generation data, and task-specific enhancement data, with rejection sampling retained to filter noisy samples and further mitigate hallucinations.

This stage boosts the model’s performance on specific tasks while maintaining the balance between understanding and generation, ensuring practical utility in real-world scenarios.

\begin{table}[t!]
\centering
\footnotesize 
\renewcommand{\arraystretch}{1.4} 
\setlength{\tabcolsep}{2pt} 
\begin{tabular}{p{4cm}ccp{9cm}} 
\toprule
Stage & Samples & Tokens & Data Type \\
\midrule
Stage 1:\newline Cross-Modal Alignment & 191M & 108B & Pure text corpora; \newline image-caption pairs  
\\
\midrule
Stage 2.1:\newline Interleaved Scaling & 173M & 191B & 
Large-scale interleaved image-text data \\
Stage 2.2:\newline Multi-Task Pre-Training & 593M & 679B & Large-scale pure text corpora; \newline descriptive captions; simple visual-language QA; \newline text-to-image pairs (reversed from captions) \\
\midrule
Stage 3:\newline Annealing Training & 40M & 70B & Pure text Chain-of-Thought (CoT) data; \newline multimodal CoT data; high-quality filtered multimodal QA; \newline filtered pure text corpora; general image generation data (filtered via rejection sampling) \\
\midrule
Stage 4:\newline SFT & 4M & 8B & Replayed Stage 3 high-quality data; \newline task-specific CoT data; task-specific image generation data (e.g., attribute/subject/scene/style/perspective control); \newline task-specific enhancement data (\textit{e.g.}, complex OCR, math reasoning, STEM data) \\
\bottomrule
\end{tabular}
\caption{Data statistics of each LLM training stage, including the number of samples, total token volume, and data type.}
\label{tab:ar_training_data}
\end{table}

\subsection{Kelix-DiT Training Recipes}
We exclusively target 1024×1024 high-resolution image generation and adopt a two-stage training paradigm, which balances the model’s generalization ability across diverse scenarios and fine-grained adherence to textual instructions.

\subsubsection{Training Setting}
Kelix-DiT is conditioned on all last hidden states between the vision start token and the vision end token.
All DiT parameters are updated during training.

\subsubsection{Two-Stage Training Strategy}
\paragraph{Pretraining Stage} The pretraining dataset is constructed by inverting large-scale image-caption pairs, with source images constrained to an aspect ratio of 0.67–1.5. These images are then resized to 504×504 for condition extraction and 1024×1024 for target construction. The dataset’s diversity in scenes, object categories, and image scales equips the model with robust semantic-image alignment capabilities and strong generalization to unseen scenarios.

\paragraph{SFT Stage} We use high-quality text-to-image data filtered via rejection sampling to enhance the model’s instruction-following capability. The dataset is carefully curated to cover key fine-grained control dimensions, including attribute control, subject control and so on. The target image is strictly constrained to a resolution of 1024x1024.

\section{Evaluation}

\subsection{Benchmarks}
In this section, we evaluate our model’s image understanding and generation capabilities across diverse benchmarks, covering core task scenarios and ensuring rigorous assessment.

\subsubsection{Image Understanding Benchmarks}
We categorize the evaluation into three types of benchmarks, targeting distinct visual-linguistic reasoning abilities:
\begin{itemize}
    \item \textbf{General VQA Benchmarks}: %
    RealworldQA~\citep{grok} focuses on daily visual question answering with diverse scenes and question types; MMBench$_\text{EN}$~\citep{liu2024mmbench} is an English multimodal benchmark covering various visual-linguistic tasks; SEED-Bench~\citep{li2023seed} consists of 19K human-annotated multiple-choice questions spanning 12 evaluation dimensions including image understanding and generative comprehension. %
    \item \textbf{Real-World Reasoning Benchmarks}: %
    AI2D~\citep{kembhavi2016diagram} specializes in multi-step reasoning for diagram interpretation; MMMU~\citep{yue2024mmmu} is an interdisciplinary multimodal understanding and reasoning benchmark covering professional knowledge across domains; MathVista~\citep{lu2023mathvista} evaluates mathematical reasoning capabilities in visual contexts. %
    \item \textbf{Text-Rich Benchmarks}: %
    ChartQA~\citep{liu2024chatqa} focuses on textual semantic understanding and question answering for chart data; TextVQA~\citep{singh2019towards} tests the ability to recognize and understand scene text; OCRbench~\citep{Liu_2024} comprehensively assesses model performance on OCR-related tasks. With unambiguous task definitions and consistent annotations, these benchmarks exhibit the highest stability and serve as the primary basis for verifying core design effectiveness.
\end{itemize}

\subsubsection{Image Generation Benchmarks}
We select three complementary benchmarks to comprehensively evaluate text-to-image alignment, knowledge grounding, and complex semantic modeling:
\begin{itemize}
    \item \textbf{GenEval}~\citep{ghosh2023geneval}: An object-focused framework assessing fine-grained text-image alignment (\textit{e.g.}, count, color, position, attributes) with high human annotation agreement, distinguishing simple/complex prompt performance.
    \item \textbf{WISE}~\citep{niu2025wise}: A world knowledge-informed benchmark verifying if generated images comply with common sense (\textit{e.g.}, cultural norms, physical laws, biology), focusing on knowledge grounding.
    \item \textbf{DPG-Bench}~\citep{hu2024ella}: Evaluates complex semantic alignment and instruction-following via dense structured prompts, testing entity recognition, attribute modeling, and relational reasoning with ``hard prompts'' for model differentiation.
\end{itemize}

\subsection{Image Understanding Performance} 
We evaluate the understanding performance of Kelix on diverse perspectives in Table \ref{tab:full_comprehensive_evaluation_clean}. Our Kelix (8B) showcases clear advantages when compared to other unified multimodal LLMs, regardless of whether they employ discrete, continuous, or hybrid tokenization schemes. 

Among discrete-tokenization unified models, Kelix achieves a decisive performance edge. For instance, it outperforms Janus-Pro (7B) by a substantial margin of 28.4 points on OCRBench (86.7 vs. 59.0) and surpasses Emu3 (8B) by 29.0 points on MathVista (76.5 vs. 47.6), demonstrating remarkable improvements on both text-rich and reasoning-intensive tasks.

Notably, Kelix narrows or even eliminates the long-standing performance gap with models built on continuous or hybrid tokenization, which are conventionally regarded as possessing superior visual representational fidelity. Specifically, Kelix attains a MathVista score of 76.5, surpassing the 14B Bagel (73.1) and the 30B Manzano (73.3) despite its considerably more compact parameter scale. Furthermore, in text-rich scenarios, Kelix attains state-of-the-art performance among all open-source unified multimodal models; its OCRBench score of 86.7 even matches or surpasses that of proprietary models such as GPT-4o (73.6) and Gemini-2.5-Pro (86.2). These robust results solidly establish Kelix as a powerful and versatile unified framework for general multimodal understanding.

\paragraph{Kelix overcomes the information bottleneck.}
Furthermore, existing discrete-tokenization unified models typically suffer substantial performance degradation on text-rich benchmarks, a critical limitation stemming from the inherent information bottleneck of single-token quantization approaches, as discussed in Sec. ~\ref{sec:discrete_image_tokenizer}. For instance, X-Omni (7B), the second-best discrete-tokenization baseline, achieves an OCRBench score of 70.4 and a TextVQA score of 77.4, which are notably lower than understanding-only Qwen2.5-VL (7B) (86.4 on OCRBench and 84.9 on TextVQA). %

By contrast, Kelix stands as the first discrete-tokenization unified model to achieve OCRBench performace on par with that of understanding-only VLMs. Specifically, Kelix attains an OCRBench score of 86.7, which is nearly identical to the 86.4 achieved by VLM Qwen2.5-VL (7B), while outperforming the second-best discrete-tokenization baseline X-Omni (7B) by a substantial 16.3 points, corresponding to a relative performance improvement of 23\%. The result validates that Kelix’s product quantization mechanism effectively bridges the representational gap between continuous visual embeddings and discrete tokens, demonstrating that discrete visual representations are no longer a fundamental performance bottleneck for multimodal understanding even for text-rich tasks that demand fine-grained semantic preservation.

\subsection{Image Generation Performance}

Tables \ref{tab:generation_results} and \ref{tab:dpg_bench_results} present the image generation performance of Kelix against competitive baselines across representative benchmarks. 

On the GenEval benchmark, Kelix (8B) attains a state-of-the-art (SOTA) score of 87.6 for discrete-tokenization unified models, a result that significantly outperforms the second-ranked models in this paradigm, Nextflow (7B, 83.0) and X-Omni (7B, 83.0). Notably, this score even exceeds that of the large-scale Qwen-Image (27B), which reaches 87.0, by a 0.6-point margin. When extending the comparison to all models across continuous, hybrid, and discrete tokenization paradigms, Kelix’s performance is second only to TUNA (7B, 90.0). Specifically, Kelix outperforms the second best discrete-tokenization unified models by substantial margins across several sub-tasks: 6 points in Counting, 6 points in Position, and 10 points in Color attributes. %

For the WISE benchmark, which evaluates world knowledge grounding in generated images, Kelix attains a score of 57.0, ranking second only to Nextflow (7B, 59.0). Critically, Kelix outperforms all continuous-tokenization-based unified models, including the larger-scale Bagel (14B, 52.0), as well as dedicated text-to-image (T2I) models such as the larger FLUX.1-dev (12B, 50.0). On DPG-Bench, which measures compositional generation and complex semantic alignment capabilities, Kelix achieves a score of 85.5, which is highly competitive with the SOTA performance of X-Omni (7B, 87.7) with only a slight 2.2-point gap. The result confirms the robust capabilities of Kelix in image generation and fine-grained text-image alignment.

Notably, Kelix is trained without the reinforcement learning stage adopted in some other baselines, e.g., X-Omni, which further validates that our framework can strike a good balance between multimodal understanding and high-quality image generation capabilities within a fully discrete autoregressive architecture.

\section{Ablation Experiments}
\subsection{Ablation Study on Kelix-Tok}

\definecolor{colA}{RGB}{230, 242, 255} 
\definecolor{colB}{RGB}{230, 250, 245} 
\definecolor{colC}{RGB}{255, 248, 220} 
\definecolor{colD}{RGB}{255, 235, 225} 
\definecolor{colE}{RGB}{255, 230, 240} 
\definecolor{colF}{RGB}{240, 235, 255} 
\definecolor{colG}{RGB}{240, 240, 240} 

\begin{table*}[t]
    \centering
    \scriptsize 
    \setlength{\tabcolsep}{2.5pt} 
    \renewcommand{\arraystretch}{1.25} 
    
    \caption{
        Ablation studies on \textbf{Kelix-Tok}. 
    }
    \label{tab:ablation_MT_tokenizer}
    
    \begin{tabular}{
        @{} l  
        cccc       
        cccc       
        cccc       
        @{}
    }
        \toprule
        \multirow{2.5}{*}{\textbf{Configuration}} & 
        \multicolumn{4}{c}{\textbf{General Benchmarks}} & 
        \multicolumn{4}{c}{\textbf{Knowledge Benchmarks}} & 
        \multicolumn{4}{c}{\textbf{Text-rich Benchmarks}} \\
        
        \cmidrule(r){2-5} \cmidrule(lr){6-9} \cmidrule(l){10-13}
        
        & RWQA & MMB$_\text{CN}$ & MMB$_\text{EN}$ & \multicolumn{1}{c}{\textbf{Avg.}} 
        & AI2D & MMMU & MathV & \multicolumn{1}{c}{\textbf{Avg.}}
        & Chart & Text & OCRB & \multicolumn{1}{c}{\textbf{Avg.}} \\
        \midrule
        
        \multicolumn{13}{c}{\textit{(a) Baseline Comparison}} \\
        \cdashline{1-13}
        \rowcolor{colA}
        Discrete Baseline
        & 55.3 & 53.8 & 57.8 & 55.6
        & 61.6 & 31.3 & 40.3 & 44.4
        & 68.7 & 64.8 & 66.0 & 66.5 \\
        \rowcolor{colA}
        Continuous Baseline
        & 59.7 & 56.6 & 63.1 & 59.8
        & 65.4 & 34.2 & 50.3 & 50.0
        & 78.9 & 75.8 & 79.3 & 78.0 \\
        \rowcolor{colA}
        Kelix-Tok (Default)
        & 58.8 & 55.9 & 59.3 & 58.0
        & 62.9 & 33.9 & 50.0 & 48.9
        & 76.5 & 74.7 & 79.2 & 76.8 \\
        
        \specialrule{\lightrulewidth}{0pt}{0pt}

        \multicolumn{13}{c}{\textit{(b) Independent vs. Shared Codebooks (fixed $N=8$, Mean Pool)}} \\
        \cdashline{1-13}
        \rowcolor{colC}
        Shared Codebook
        & 49.0 & 46.8 & 50.6 & 48.8
        & 59.0 & 34.8 & 44.3 & 46.0
        & 67.0 & 60.4 & 61.7 & 63.0 \\
        \rowcolor{colC}
        Independent (Default)
        & 57.4 & 56.3 & 59.3 & 57.7
        & 64.6 & 32.4 & 48.3 & 48.4
        & 76.4 & 73.3 & 76.5 & 75.4 \\
        
        \specialrule{\lightrulewidth}{0pt}{0pt}

        \multicolumn{13}{c}{\textit{(c) Fusion Strategy (fixed $N=8$)}} \\
        \cdashline{1-13}
        \rowcolor{colD}
        Mean Pooling
        & 57.4 & 56.3 & 59.3 & 57.7
        & 64.6 & 32.4 & 48.3 & 48.4
        & 76.4 & 73.3 & 76.5 & 75.4 \\
        \rowcolor{colD}
        Sum Pooling (Default)
        & 58.8 & 55.9 & 59.3 & 58.0
        & 62.9 & 33.9 & 50.0 & 48.9
        & 76.5 & 74.7 & 78.5 & 76.6 \\
        
        \specialrule{\lightrulewidth}{0pt}{0pt}

        \multicolumn{13}{c}{\textit{(d) Codebook Dimension (d)}} \\
        \cdashline{1-13}
        \rowcolor{colE}
        $d=32$
        & 54.9 & 49.4 & 54.6 & 53.0
        & 60.1 & 30.7 & 43.3 & 44.7
        & 69.1 & 62.4 & 64.0 & 65.2 \\
        \rowcolor{colE}
        $d=64$
        & 53.3 & 52.6 & 56.6 & 54.2
        & 61.0 & 32.0 & 42.2 & 45.1
        & 69.4 & 65.4 & 65.8 & 66.9 \\
        \rowcolor{colE}
        $d=128$ (Default)
        & 55.3 & 53.8 & 57.8 & 55.6
        & 61.6 & 31.3 & 40.3 & 44.4
        & 68.7 & 64.8 & 66.0 & 66.5 \\
        \rowcolor{colE}
        $d=256$
        & 52.0 & 50.9 & 51.9 & 51.6
        & 59.8 & 29.0 & 44.1 & 44.3
        & 71.1 & 63.3 & 63.8 & 66.1 \\
        \rowcolor{colE}
        $d=512$
        & 50.6 & 52.9 & 56.3 & 53.3
        & 59.9 & 29.6 & 44.8 & 44.8
        & 69.8 & 66.6 & 66.7 & 67.7 \\
        \rowcolor{colE}
        $d=1024$
        & 49.5 & 52.6 & 54.2 & 52.1
        & 60.1 & 32.0 & 41.7 & 44.6
        & 68.0 & 66.3 & 64.4 & 66.2 \\
        
        \specialrule{\lightrulewidth}{0pt}{0pt} 
        
        \multicolumn{13}{c}{\textit{(e) Codebook Size (\textit{S})}} \\
        \cdashline{1-13}
        \rowcolor{colF}
        \textit{S}=32,768
        & 54.0 & 50.1 & 53.9 & 52.7
        & 60.3 & 31.0 & 42.6 & 44.6
        & 70.0 & 64.0 & 65.0 & 66.3 \\
        \rowcolor{colF}
        \textit{S}=65,536 (Default)
        & 55.3 & 53.8 & 57.8 & 55.6
        & 61.6 & 31.3 & 40.3 & 44.4
        & 68.7 & 64.8 & 66.0 & 66.5 \\
        \rowcolor{colF}
        \textit{S}=131,072
        & 54.0 & 48.4 & 51.9 & 51.4
        & 59.6 & 32.6 & 43.8 & 45.3
        & 70.2 & 65.1 & 65.9 & 67.1 \\
        \bottomrule
    \end{tabular}
\end{table*}

In this section, we conduct a comprehensive set of ablation studies to systematically evaluate the impact of key design decisions in the Kelix-Tok. Table~\ref{tab:ablation_MT_tokenizer} summarizes our results. For efficiency, all model variants are trained on a representative subset of the full dataset and only adopt the two-stage training protocol (Stage 1 and Stage 3) — distinct from the main experiment’s complete training workflow. Each experiment follows a single-factor ablation protocol: starting from the Discrete Baseline, we modify one component at a time to precisely isolate its effect on performance.

\textbf{Baseline Comparison.} The results in Table~\ref{tab:ablation_MT_tokenizer}(a) reveals a critical trade-off among different visual representation schemes. We observe that, compared to the Continuous Baseline which represents a performance upper bound, the Discrete Baseline introduces a significant performance gap. This is particularly pronounced on Text-rich tasks, where its performance drops to approximately 85\% of the continuous baseline (66.5 vs. 78.0 Avg.) due to the information bottleneck. 
Our Kelix-Tok, leveraging a PQ-based combinatorial representation, successfully achieves discretization while attaining performance comparable to its continuous counterpart.

\textbf{Independent vs. Shared Codebooks.}
We validate the necessity of our independent sub-codebook design. As shown in Table~\ref{tab:ablation_MT_tokenizer}(b), forcing all sub-tokens to a single Shared Codebook, as opposed to our default Independent codebooks, result in a catastrophic decline in performance (e.g., -12.3\% on average for Text-rich tasks). This finding highlights that sub-codebooks are essential for learning disentangled and meaningful visual concepts; utilizing a shared semantic space induces severe representation confusion.  These results underscore the critical role of our independent codebook strategy in achieving effective visual tokenization.

\textbf{Fusion Strategy.}
With $N=8$, we compare Mean Pooling and Sum Pooling as strategies for fusing sub-token embeddings. As shown in Table~\ref{tab:ablation_MT_tokenizer}(c), Sum Pooling consistently achieves equal or superior performance across all tasks, with a notable advantage on detail-sensitive Text-rich benchmarks (76.6 vs. 75.4 Avg.). This supports our hypothesis that summation more effectively preserves the overall information content and signal magnitude of the combinatorial representation than averaging, thus more effectively conveying visual details to the language model. Consequently, we adopt Sum Pooling as our default fusion strategy.

\textbf{Codebook Dimension.}
As shown in Table~\ref{tab:ablation_MT_tokenizer}(d), performance peaks at $d=128$. Smaller dimensions ($d \le 64$) lead to performance degradation due to an information bottleneck, while larger dimensions ($d \ge 256$) likely hinder optimization by introducing excessive sparsity in the representation space. Therefore, $d=128$ provides an optimal balance between representational capacity and optimization tractability, and is adopted as our default setting.

\textbf{Codebook Size.}
Finally, we investigate the impact of codebook size on the discrete baseline. Results in Table~\ref{tab:ablation_MT_tokenizer}(e) show that increasing the size($S$) from 32K to 64K yields consistent gains. However, a further increase to 128K leads to performance degradation, likely due to insufficient training of an overly large codebook with limited data (i.e., codebook under-utilization). We thus select 65,536 as the optimal size, balancing representational capacity with training efficiency.

\textbf{Quantization Strategies.}
To investigate the impact of different discretization schemes, we compare \textbf{(1)} \textbf{Vector Quantization (VQ)} ~\citep{van2017neural,zhu2025addressing} with a codebook size of 65536, \textbf{(2)} \textbf{Finite Scalar Quantization (FSQ)}~\citep{mentzer2023finite} with an equivalent index space ($16^4 = 65536$), and \textbf{(3)} \textbf{Residual Quantization (RQ)}~\citep{lee2022autoregressive} utilizing a three-layer hierarchical structure ($3 \times 8192$). 
OCRBench is employed to evaluate their performance in fine-grained text recognition and document understanding. 
We observe that \textbf{VQ achieves the highest score of 66.0}, outperforming RQ (60.3) and FSQ (49.2). This performance gap indicates that the learnable embeddings in VQ offer superior representational flexibility for capturing dense visual-textual features. 
Consequently, VQ is adopted as the default quantization strategy for our final model.

\captionsetup[subfigure]{justification=centering}  
\captionsetup[figure]{justification=centering}     

\begin{figure}[t!]
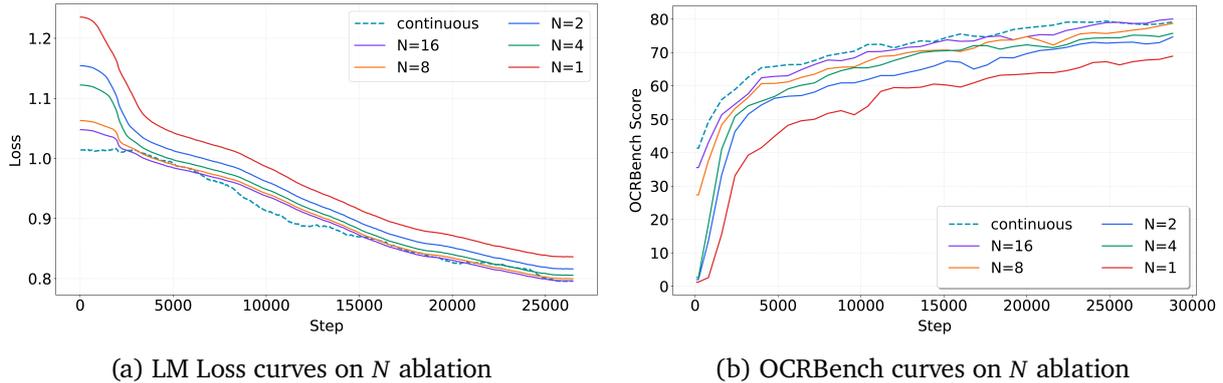

  \centering  
  \begin{subfigure}{0.4775\textwidth}
    \centering  
    \includegraphics[width=\textwidth]{figures/stage3_loss_comparison.png}
    \subcaption{LM Loss curves on $N$ ablation}  
    \label{fig:N-ab-loss}
  \end{subfigure}
  \hspace{0.5mm}  
  \begin{subfigure}{0.475\textwidth}
    \centering  
    \includegraphics[width=\textwidth]{figures/stage3_ocr_bench_score_curve.png}
    \subcaption{OCRBench curves on $N$ ablation}
    \label{fig:N-ab-ocr}
  \end{subfigure}
  
  \caption{Ablation study on parameter $N$: LM Loss and OCRBench curves.}
  \label{fig:N-ab}
\end{figure}

\textbf{Number of Sub-Tokens.} We investigate the impact of varying the number of sub-tokens $N$ by conducting experiments with $N=1,2,4,8,16$ alongside a continuous baseline, with results presented in Fig.~\ref{fig:N-ab}. As shown in Subfigure.~\ref{fig:N-ab-loss}, the LM loss consistently decreases with increasing $N$, and the loss curves for $N \geq 8$ converge closely to the continuous baseline, indicating that sufficient discrete tokens can effectively compress continuous visual information without substantial loss. We further evaluate our training checkpoints on the OCRBench benchmark, which are highly sensitive to model performance.  Subfigure.~\ref{fig:N-ab-ocr} demonstrates that OCRBench scores consistently increase with $N$. Upon training convergence, the performance of the continuous baseline is nearly matched by the top two discrete configurations ($N=8$ and $N=16$), which further validates the robustness of our PQ-based Kelix.

\subsection{Ablation Study on Kelix-LLM}
\begin{table*}[t]
    \centering
    \scriptsize 
    \setlength{\tabcolsep}{2.5pt} 
    \renewcommand{\arraystretch}{1.25} 
    
    \caption{
        Ablation studies on our \textbf{Unified Multimodal LLM Architecture}. 
        We evaluate the design of multimodal interfaces, representation paradigms, and training synergies across Stage 1 and Stage 3. 
    }
    \label{tab:ablation_unified_model}
    
    \begin{tabular}{
        @{} l  
        cccc       
        cccc       
        cccc       
        @{}
    }
        \toprule
        \multirow{2.5}{*}{\textbf{Configuration}} & 
        \multicolumn{4}{c}{\textbf{General Benchmarks}} & 
        \multicolumn{4}{c}{\textbf{Knowledge Benchmarks}} & 
        \multicolumn{4}{c}{\textbf{Text-rich Benchmarks}} \\
        
        \cmidrule(r){2-5} \cmidrule(lr){6-9} \cmidrule(l){10-13}
        
        & RWQA & MMB$_\text{CN}$ & MMB$_\text{EN}$ & \multicolumn{1}{c}{\textbf{Avg.}} 
        & AI2D & MMMU & MathV & \multicolumn{1}{c}{\textbf{Avg.}}
        & Chart & Text & OCRB & \multicolumn{1}{c}{\textbf{Avg.}} \\
        \midrule
        
        %
        exp1: Unified Baseline
        & 61.3 & 67.7 & 66.9 & 65.3 &
        70.8	& 42.0 &	55.8 &	56.2 &
        76.9 &	72.3	& 79.9 &	76.4 \\
        \rowcolor{colA}
        exp2: Random Init Input Emb.
        & 56.7 &	60.7&	59.3&	58.9	
        &67.5&	34.7&	53.8&	52.0	
        &77.8	&70.4&	77.0 &	75.1 \\
        \rowcolor{colA}
        exp3: 9-Head Predictor (Output)
        & 64.8	&60.4	&61.6	&62.3 
        &	69.3	&41.7&	52.9 &	54.6 
        &	75.2 &	71.4 &	76.5 &	74.4  \\

        \bottomrule
    \end{tabular}
\end{table*}

In this section, we ablate the key design components in our unified LLM architecture, including the multimodal input/output interface, representation paradigm, and training data synergy, to validate their effectiveness. The experimental setting are aligned with the settings in Table~\ref{tab:ablation_unified_model}, where the exp1 served as the baseline and modify one core component at a time to isolate its impact on model performance. In the following, we detail the training setting and analyze the the experimental results, and derive insights for the unified LLM framework.

\paragraph{Experimental Setting Explanations.}

All ablation variants are built on the Qwen3-8B backbone but only adopt a two-stage training protocol (Stage 1 and Stage 3) with a subset of full training data, unlike the full four-stage workflow of the main experiment. This simplification streamlines the ablation process by eliminating redundant training stages that are unnecessary for isolating the effects of target components. It ensures the ablation focuses on core design validations without excessive computational overhead. Key ablations are defined as follows:
\begin{itemize}

\item \textbf{exp1: Unified Baseline}: The full-version of our proposed unified LLM model. It includes: (1) Stage 1 training with a Linear Aligner to map VQ discrete tokens to the LLM's semantic space; (2) a unified \textit{Block Encoder} (sum pooling) and \textit{Block Decoder} for multimodal input aggregation and output reconstruction; (3) training on both understanding-oriented datasets (e.g., VQA, OCR) and generation-oriented datasets (e.g., text-to-image pairs). This model variant serves as the performance benchmark for all ablations.

\item \textbf{exp2: Random Init Input Emb.}: Ablates the \textit{up-projector} component in LLM Stage 1. Starting from LLM Stage 1, we directly use token IDs from the VQ quantizer as visual inputs, and initialize the visual embedding table randomly without any alignment training. This setting could test whether the Linear Aligner and Stage 1's cross-modal alignment strategy are necessary.

\item \textbf{exp3: 9-Head Predictor (Output)}: Replaces the \textit{Block Decoder} with a 9-head parallel predictor. Specifically, the LLM's single hidden state is projected into 9 independent vectors via 9 separate linear heads, which are then fed into the LM head to generate 9 discrete tokens. For text generation, only the first token is retained; for image generation, the latter 8 tokens are used. This variant verifies the rationality of the \textit{Block Decoder}'s hierarchical reconstruction mechanism.

\end{itemize}

\paragraph{Result Analysis}

We analyze the ablation results across three benchmark categories (General Benchmarks, Knowledge Benchmarks, Text-rich Benchmarks) and focus on key performance differences between variants.

The results in Table~\ref{tab:ablation_unified_model} highlight the critical role of the \textit{up-projector} and \textit{Block Encoder} in cross-modal modeling:

\begin{itemize}

\item \textbf{Effect of Linear Aligner (exp1 vs. exp2)}: Compared to the baseline (exp1, General Avg. 65.3), the variant using randomly initialized visual embedding table (exp2) shows a significant performance drop across all tasks: General Avg. decreases by 6.4 points (to 58.9), Knowledge Avg. drops by 4.2 points (56.2 → 52.0), and Text-rich Avg. decrease by 1.3 points (76.4 → 75.1). This confirms that Stage 1's \textit{up-projector} effectively bridges the semantic gap between discrete visual tokens and the LLM's text space, avoiding catastrophic forgetting of linguistic priors and laying the foundation for cross-modal reasoning. Without this alignment, random initialization of visual embeddings leads to misalignment between modalities, severely degrading model performance.

\item \textbf{Effect of \textit{Block Encoder} (exp1 vs. exp3) }: Replacing the \textit{Block Encoder} with a 9-head predictor (exp3) results in consistent performance degradation: General Avg. decreases by 3.0 points (65.3 → 62.3), Knowledge Avg. drops by 1.6 points (56.2 → 54.6), and Text-rich Avg. falls by 2.0 points (76.4 → 74.4). The 9-head design treats text and image outputs as independent token streams, breaking the semantic integrity of "blocks" (text blocks as single semantic units, image blocks as 8-token spatial clusters). This indicates that the \textit{Block Encoder}'s hierarchical reconstruction (aggregating hidden states into blocks and then reconstructing tokens) better preserves the structural and semantic consistency of multimodal data. In contrast, the parallel 9-head predictor introduces semantic fragmentation, especially for image tokens that require spatial coherence, thus reducing model performance.

\begin{figure}[t!]
  \centering
  \includegraphics[width=1\textwidth]{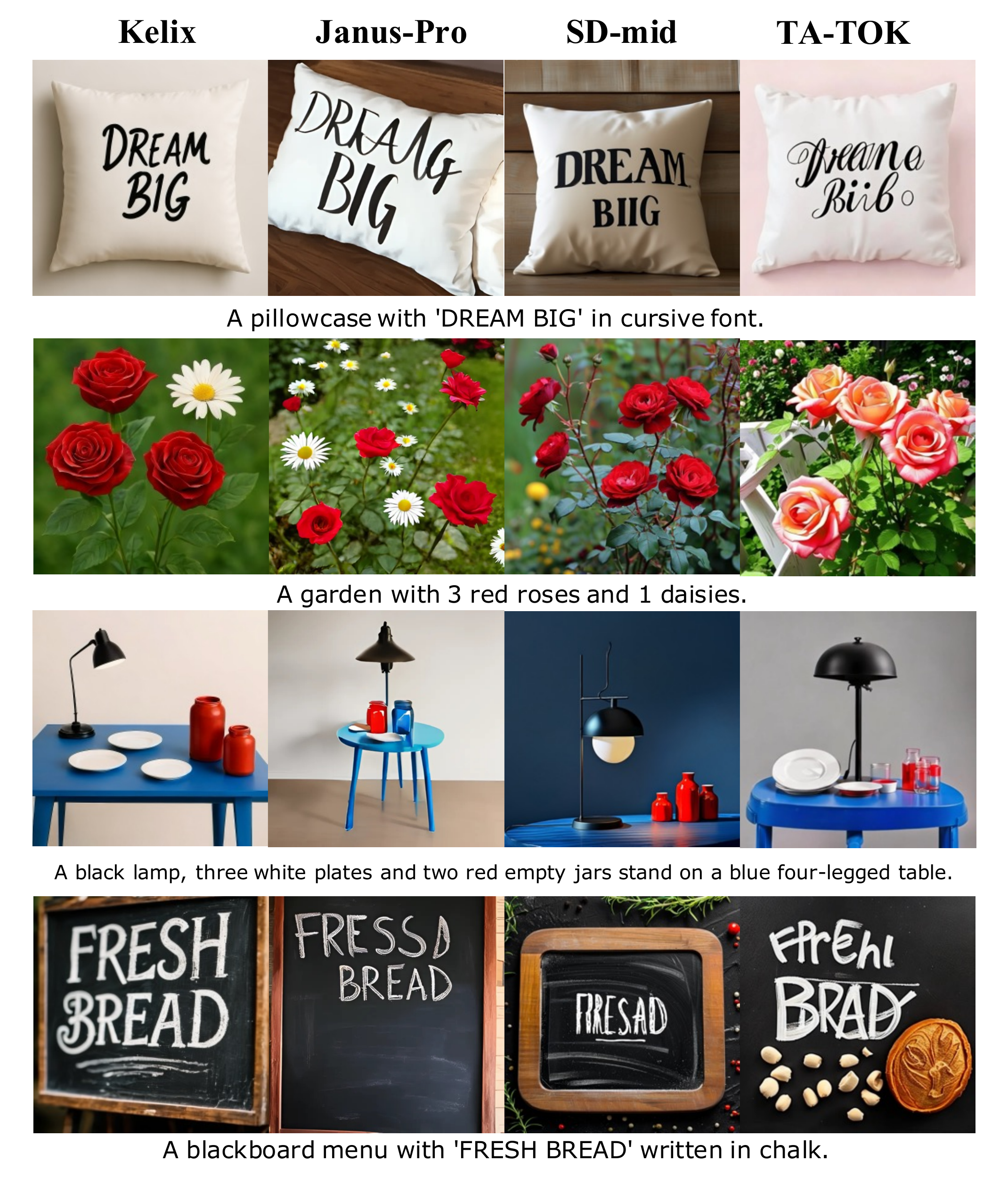}
    \caption{The visualization comparison of different unified models in image generation.}
    \label{fig:casecompare}
\end{figure}

\end{itemize}

\paragraph{Key Findings}

From the ablation study on the unified LLM architecture, we draw two critical conclusions:

\begin{enumerate}

\item \textbf{Cross-modal alignment is indispensable}: The \textit{up-projector} in Stage 1 effectively maps discrete visual tokens to the LLM's semantic space, avoiding inter-modal misalignment and ensuring the model's basic cross-modal reasoning ability.

\item \textbf{Unified block-wise processing is superior}: The \textit{Block Decoder}'s hierarchical reconstruction mechanism preserves the semantic integrity of multimodal data, outperforming parallel multi-head predictors that cause semantic fragmentation.

\end{enumerate}

These findings confirm the rationality of our unified LLM architecture design, providing empirical support for the core components (the up-projector and Block Embedding Moduling) and laying the foundation for the model's superior performance in both understanding and generation tasks.

\subsection{Case Study of Image Generation}

We show image generation cases of Kelix and other SOTA unified models in Figure~\ref{fig:casecompare}. The visualization results demonstrate that Kelix achieves superior instruction-following capability compared to baselines.
In the first case (pillowcase with ``DREAM BIG'' in cursive), the Janus-Pro, SD-mid, and TA-TOK produces either unclear or non-cursive text.
In the second case (garden with 3 red roses and 1 daisy), they generate incorrect flower counts or mix up the flower types.
In the third case (one black lamp, three white plates, and two red jars, a four-legged table), all of these baseline methods do not incorrectly render the number of plates or jars.
These comparisons highlight that Kelix better adheres to fine-grained instructions including text rendering, object counting, and spatial arrangement.

\section{Related Works}

We review prior work along the two representation dimensions introduced in Section~\ref{sec2_Design}: \emph{representation form} (discrete vs.\ continuous) and \emph{semantic level} (high-level vs.\ low-level), covering vision-language models, unified multimodal models, and image generation techniques.

\subsection{Vision-Language Models}

CLIP~\citep{radford2021learning} established the foundation for modern VLMs by aligning visual and textual semantics in a shared embedding space via large-scale contrastive pre-training. Building on CLIP's pre-trained ViT, Flamingo~\citep{alayrac2022flamingo} introduced cross-attention mechanisms to fuse visual and language representations, enabling few-shot visual reasoning. However, the cross-attention design modifies the LLM architecture, limiting compatibility with pre-trained LLM backbones.

BLIP-2~\citep{li2023blip} addressed this by proposing a lightweight Q-Former that bridges a frozen ViT and a frozen LLM, requiring only the adaptor to be trained. LLaVA~\citep{liu2023visual} further simplified the pipeline to a ``ViT--Projector--LLM'' framework, feeding fine-grained patch embeddings directly to the LLM. This paradigm has since become the dominant VLM design, adopted by Qwen2-VL, Qwen2.5-VL~\citep{wang2024qwen2, bai2025qwen2}, MiniCPM-V~\citep{yao2024minicpm}, InternVL~\citep{chen2024internvl}, among others.

More recently, VLMs have shifted toward native-resolution ViT encoders to process images at their original resolution without complex splitting or resizing, enabling more fine-grained visual understanding. Representative models include Qwen3-VL~\citep{Qwen3-VL}, Kimi-VL~\citep{team2025kimi}, Seed-VL -1.5~\citep{guo2025seed1}, and Keye-VL~\citep{yang2025kwai}. Note that all these VLMs rely on \emph{continuous} visual features and are limited to understanding tasks.

\subsection{Unified Multimodal Models}

Although understanding-only VLMs have achieved great success, the absence of generation capability limits both the application scope and the scaling potential of these models. This has motivated the community to develop unified models that support both understanding and generation. Existing approaches broadly fall into two categories: \emph{unified architectures} and \emph{hybrid architectures}.

\textbf{Unified Architectures.}
In unified architectures, a single representation serves both understanding and generation, and discrete tokenization is a natural choice for this paradigm. Early works such as Chameleon~\citep{team2024chameleon}, LlamaGen~\citep{sun2024autoregressive}, and Emu3~\citep{wang2024emu3} adopt VQ-VAE-like techniques~\citep{van2017neural} to encode images into low-level discrete codes, achieving unified text-vision modeling under next-token prediction. However, the semantic level mismatch between low-level visual codes and high-level text tokens leads to low cross-modal alignment efficiency and poor sample efficiency during scaling.

More recently, a growing number of works have turned to high-level visual semantics derived from ViT encoders, including BLIP3o~\citep{chen2025blip3ofamilyfullyopen}, BLIP3o-NeXT~\citep{chen2025blip3o}, TaTok~\citep{han2025vision}, X-Omni~\citep{geng2025xomnireinforcementlearningmakes}, UniTok~\citep{ma2025unitok}, and NextFlow~\citep{zhang2026nextflow}. High-level semantics exhibit significantly higher alignment efficiency, enabling fast transfer of the LLM's world knowledge and reasoning capabilities to visual generation, and these models have achieved strong results on generation benchmarks. However, discrete tokens in this route still face a fundamental challenge: the information loss inherent in discretization causes them to perform significantly worse than contemporary understanding-only VLMs on multimodal understanding tasks, which to some extent hinders the realization of true modal unification under this framework.

\textbf{Hybrid Architectures.}
To address the understanding weakness of unified architectures, hybrid models adopt separate representations for understanding and generation. On the understanding side, they retain the proven ViT--Projector--LLM pipeline from understanding-only VLMs; on the generation side, they employ a different representation for visual synthesis. Early hybrid models use low-level discrete codes for generation, such as Janus~\citep{wu2025janus}, Janus-Pro~\citep{janusPro} (VQ-VAE-based) and UniHetero~\citep{chen2025unihetero}, or low-level continuous features, such as JanusFlow~\citep{ma2025janusflow} and BAGEL~\citep{deng2025emerging}. Following the same trend as unified architectures, recent hybrid works have also begun exploring high-level semantics for the generation branch, including MetaQuery~\citep{pan2025transfer}, and Pisces~\citep{xu2025pisces}

The fundamental limitation of hybrid architectures is the representational asymmetry between understanding and generation, which inherently impairs knowledge sharing and cross-task transfer. Recent efforts such as Manzano~\citep{li2025manzano}, TuNa~\citep{liu2025tuna}, Ming-UniVision~\citep{mingUniVision} and RAE~\citep{tong2026scaling} have attempted to address this by unifying the representation space into high-level semantics, achieving some positive results.

\textbf{Our Position.}
Our proposed \myorange{\textbf{Kelix}} follows the \textbf{unified architecture} route with \textbf{high-level discrete semantics}, but introduces a \myorange{product quantization mechanism} that expands the representational capacity of discrete tokens without increasing the LLM's sequence length. This directly addresses the information loss bottleneck of discrete tokenization, aiming to close the understanding gap between discrete and continuous representations and unlock the full potential of unified multimodal modeling.

\subsection{Image Generation}

Image Generation is undergoing a profound transformation. The core architecture of diffusion models is shifting from convolutional U-Net backbones~\citep{rombach2022high} toward more scalable and parameter-efficient Diffusion Transformers (DiT)~\citep{peebles2023scalable, team2025zimage}, with training objectives evolving from traditional noise prediction to more stable and expressive flow-matching paradigms~\citep{ma2025janusflow, xie2025sana}. A pivotal driver behind this evolution is the upgrade of conditioning mechanisms: transitioning from static, fixed-vocabulary text encoders like CLIP~\citep{radford2021learning} or T5 toward to more dynamic LLM-driven conditioning. As demonstrated by Qwen-Image~\citep{wang2024qwen2}, MetaQuery~\citep{pan2025transfer}, and Manzano~\citep{li2025manzano}, this paradigm harnesses the rich hidden states and world knowledge of pre-trained LLMs to guide the generative process. By treating the LLM as a sophisticated semantic engine, these systems transcend basic text-to-image mapping to unlock reasoning-heavy capabilities, such as \textbf{instruction-based editing}~\citep{qian2025pico} with nuanced spatial control and \textbf{in-context image generation}~\citep{sun2024generative} that maintains visual consistency. In this framework, image generation is effectively reimagined as a seamless extension of the LLM’s multimodal reasoning chain.

Our \textbf{Kelix} aligns with this trend but distinguishes itself by performing \textbf{semantic-level autoregressive modeling over high-level discrete tokens}. Unlike paradigms that merely utilize continuous hidden states as conditioning signals, Kelix adopts a modular \textit{Tokenizer--LLM--Detokenizer} design that fundamentally integrates generation into the LLM's next-token prediction framework~\citep{llamagen, wang2024emu3}. By predicting discrete visual representations~\citep{van2017neural}, Kelix allows the model to fully inherit the LLM’s logical capacity for complex generative tasks while leveraging a diffusion-based DiT as the \textbf{Detokenizer} to bypass the inherent quality limitations of traditional discrete decoders.

\section{Conclusion}
We present \textbf{Kelix}, a fully discrete autoregressive unified model that closes the understanding gap between discrete and continuous visual representations in multimodal models.
Our key insight is that a single discrete token cannot match the expressive capacity of a continuous embedding within a fixed vocabulary size. By decomposing each continuous patch embedding into multiple parallel discrete tokens, Kelix substantially expands the information capacity of discrete codes, enabling understanding performance comparable to continuous-feature VLMs.
To maintain efficiency despite the increased token count, we adopt a Next-Block Prediction paradigm that models grouped token blocks autoregressively, keeping the effective sequence length, and thus training and inference cost, on par with single-token approaches.
Following a modular \emph{Tokenizer--LLM--Detokenizer} design, each component can be developed and iterated independently: the tokenizer handles semantic alignment, the LLM performs reasoning and semantic control, and the diffusion-based detokenizer renders high-quality images.
Extensive experiments on major multimodal understanding and image generation benchmarks demonstrate state-of-the-art results among comparable-scale unified models, validating the effectiveness of our approach.
In the future, we plan to extend Kelix to video understanding and generation, moving toward a more general unified multimodal intelligence.

\bibliography{bibtex}

@article{radford2018improving,
  title={Improving language understanding by generative pre-training},
  author={Radford, Alec and Narasimhan, Karthik and Salimans, Tim and Sutskever, Ilya and others},
  year={2018},
  publisher={San Francisco, CA, USA}
}

@article{radford2019language,
  title={Language models are unsupervised multitask learners},
  author={Radford, Alec and Wu, Jeffrey and Child, Rewon and Luan, David and Amodei, Dario and Sutskever, Ilya and others},
  journal={OpenAI blog},
  volume={1},
  number={8},
  pages={9},
  year={2019}
}

@article{brown2020language,
  title={Language models are few-shot learners},
  author={Brown, Tom and Mann, Benjamin and Ryder, Nick and Subbiah, Melanie and Kaplan, Jared D and Dhariwal, Prafulla and Neelakantan, Arvind and Shyam, Pranav and Sastry, Girish and Askell, Amanda and others},
  journal={Advances in Neural Information Processing Systems (NeurIPS)},
  year={2020}
}

@article{hu2024ella,
  title={Ella: Equip diffusion models with llm for enhanced semantic alignment},
  author={Hu, Xiwei and Wang, Rui and Fang, Yixiao and Fu, Bin and Cheng, Pei and Yu, Gang},
  journal={arXiv preprint arXiv:2403.05135},
  year={2024}
}

@article{mentzer2023finite,
  title={Finite scalar quantization: Vq-vae made simple},
  author={Mentzer, Fabian and Minnen, David and Agustsson, Eirikur and Tschannen, Michael},
  journal={arXiv preprint arXiv:2309.15505},
  year={2023}
}

@article{li2023seed,
  title={Seed-bench: Benchmarking multimodal llms with generative comprehension},
  author={Li, Bohao and Wang, Rui and Wang, Guangzhi and Ge, Yuying and Ge, Yixiao and Shan, Ying},
  journal={arXiv preprint arXiv:2307.16125},
  year={2023}
}

@article{achiam2023gpt,
  title={Gpt-4 technical report},
  author={Achiam, Josh and Adler, Steven and Agarwal, Sandhini and Ahmad, Lama and Akkaya, Ilge and Aleman, Florencia Leoni and Almeida, Diogo and Altenschmidt, Janko and Altman, Sam and Anadkat, Shyamal and others},
  journal={arXiv preprint arXiv:2303.08774},
  year={2023}
}

@article{janusPro,
  title={Janus-pro: Unified multimodal understanding and generation with data and model scaling},
  author={Chen, Xiaokang and Wu, Zhiyu and Liu, Xingchao and Pan, Zizheng and Liu, Wen and Xie, Zhenda and Yu, Xingkai and Ruan, Chong},
  journal={arXiv preprint arXiv:2501.17811},
  year={2025}
}

@article{agarwal2025gpt,
  title={gpt-oss-120b \& gpt-oss-20b model card},
  author={Agarwal, Sandhini and Ahmad, Lama and Ai, Jason and Altman, Sam and Applebaum, Andy and Arbus, Edwin and Arora, Rahul K and Bai, Yu and Baker, Bowen and Bao, Haiming and others},
  journal={arXiv preprint arXiv:2508.10925},
  year={2025}
}

@article{kaplan2020scaling,
  title={Scaling laws for neural language models},
  author={Kaplan, Jared and McCandlish, Sam and Henighan, Tom and Brown, Tom B and Chess, Benjamin and Child, Rewon and Gray, Scott and Radford, Alec and Wu, Jeffrey and Amodei, Dario},
  journal={arXiv preprint arXiv:2001.08361},
  year={2020}
}

@article{nvidia_nemotron_nano_v3_2025,
  title={Nemotron 3 Nano: Open, Efficient Mixture-of-Experts Hybrid Mamba-Transformer Model for Agentic Reasoning},
  author={Blakeman, Aaron and Grattafiori, Aaron and Basant, Aarti and Gupta, Abhibha and Khattar, Abhinav and Renduchintala, Adi and Vavre, Aditya and Shukla, Akanksha and Bercovich, Akhiad and Ficek, Aleksander and others},
  journal={arXiv preprint arXiv:2512.20848},
  year={2025}
}

@article{karimi2025nemotroncc,
  title = {Nemotron-cc-math: A 133 Billion-Token-Scale High Quality Math Pretraining Dataset},
  author = {Rabeeh Karimi Mahabadi and Sanjeev Satheesh and Shrimai Prabhumoye and Mostofa Patwary and Mohammad Shoeybi and Bryan Catanzaro},
  journal = {arXiv preprint arXiv:2508.15096},
  year = {2025}
}

@article{nvidia2025nvidianemotronnano2,
  title={Nvidia nemotron nano 2: An accurate and efficient hybrid mamba-transformer reasoning model},
  author={Basant, Aarti and Khairnar, Abhijit and Paithankar, Abhijit and Khattar, Abhinav and Renduchintala, Adithya and Malte, Aditya and Bercovich, Akhiad and Hazare, Akshay and Rico, Alejandra and Ficek, Aleksander and others},
  journal={arXiv preprint arXiv:2508.14444},
  year={2025}
}

@misc{RJT1990_2025GeneralThoughtArchive,
      title={RJT1990/GeneralThoughtArchive},
      author={Jarius, otaldohenrikk, knight_raider and et. al.},
      year={2025},
      publisher = {Huggingface},
        howpublished = {Hugging Face Datasets},
        url={https://huggingface.co/datasets/RJT1990/GeneralThoughtArchive}
}

@misc{primeintellect2024numinamath,
  author = {PrimeIntellect},
  title = {NuminaMath-QwQ-CoT-5M},
  howpublished = {Hugging Face Datasets},
  year = {2025},
  url = {https://huggingface.co/datasets/PrimeIntellect/NuminaMath-QwQ-CoT-5M}
}

@article{ahmad2025opencodereasoning,
      title={OpenCodeReasoning: Advancing Data Distillation for Competitive Coding}, 
      author={Wasi Uddin Ahmad and Sean Narenthiran and Somshubra Majumdar and Aleksander Ficek and Siddhartha Jain and Jocelyn Huang and Vahid Noroozi and Boris Ginsburg},
      year={2025},
      eprint={2504.01943},
      archivePrefix={arXiv},
      primaryClass={cs.CL},
      url={https://arxiv.org/abs/2504.01943}
}

@article{yang2025qwen3,
  title={Qwen3 technical report},
  author={Yang, An and Li, Anfeng and Yang, Baosong and Zhang, Beichen and Hui, Binyuan and Zheng, Bo and Yu, Bowen and Gao, Chang and Huang, Chengen and Lv, Chenxu and others},
  journal={arXiv preprint arXiv:2505.09388},
  year={2025}
}

@article{Qwen3-VL,
      title={Qwen3-VL Technical Report}, 
      author={Shuai Bai and Yuxuan Cai and Ruizhe Chen and Keqin Chen and Xionghui Chen and Zesen Cheng and Lianghao Deng and Wei Ding and Chang Gao and Chunjiang Ge and Wenbin Ge and Zhifang Guo and Qidong Huang and Jie Huang and Fei Huang and Binyuan Hui and Shutong Jiang and Zhaohai Li and Mingsheng Li and Mei Li and Kaixin Li and Zicheng Lin and Junyang Lin and Xuejing Liu and Jiawei Liu and Chenglong Liu and Yang Liu and Dayiheng Liu and Shixuan Liu and Dunjie Lu and Ruilin Luo and Chenxu Lv and Rui Men and Lingchen Meng and Xuancheng Ren and Xingzhang Ren and Sibo Song and Yuchong Sun and Jun Tang and Jianhong Tu and Jianqiang Wan and Peng Wang and Pengfei Wang and Qiuyue Wang and Yuxuan Wang and Tianbao Xie and Yiheng Xu and Haiyang Xu and Jin Xu and Zhibo Yang and Mingkun Yang and Jianxin Yang and An Yang and Bowen Yu and Fei Zhang and Hang Zhang and Xi Zhang and Bo Zheng and Humen Zhong and Jingren Zhou and Fan Zhou and Jing Zhou and Yuanzhi Zhu and Ke Zhu},
	  journal={arXiv preprint arXiv:2511.21631},
      year={2025}
}

@article{team2025kimi,
  title={Kimi-vl technical report},
  author={Team, Kimi and Du, Angang and Yin, Bohong and Xing, Bowei and Qu, Bowen and Wang, Bowen and Chen, Cheng and Zhang, Chenlin and Du, Chenzhuang and Wei, Chu and others},
  journal={arXiv preprint arXiv:2504.07491},
  year={2025}
}

@article{yao2024minicpm,
  title={Minicpm-v: A gpt-4v level mllm on your phone},
  author={Yao, Yuan and Yu, Tianyu and Zhang, Ao and Wang, Chongyi and Cui, Junbo and Zhu, Hongji and Cai, Tianchi and Li, Haoyu and Zhao, Weilin and He, Zhihui and others},
  journal={arXiv preprint arXiv:2408.01800},
  year={2024}
}

@article{liu2023visual,
  title={Visual instruction tuning},
  author={Liu, Haotian and Li, Chunyuan and Wu, Qingyang and Lee, Yong Jae},
  journal={Advances in Neural Information Processing Systems (NeurIPS)},
  year={2023}
}

@article{dosovitskiy2020image,
  title={An image is worth 16x16 words: Transformers for image recognition at scale},
  author={Dosovitskiy, Alexey},
  journal={arXiv preprint arXiv:2010.11929},
  year={2020}
}

@inproceedings{peebles2023scalable,
  title={Scalable diffusion models with transformers},
  author={Peebles, William and Xie, Saining},
  booktitle={Proceedings of the IEEE/CVF International Conference on Computer Vision (ICCV)},
  year={2023}
}

@article{yang2025kwai,
  title={Kwai keye-vl 1.5 technical report},
  author={Yang, Biao and Wen, Bin and Ding, Boyang and Liu, Changyi and Chu, Chenglong and Song, Chengru and Rao, Chongling and Yi, Chuan and Li, Da and Zang, Dunju and others},
  journal={arXiv preprint arXiv:2509.01563},
  year={2025}
}

@article{van2017neural,
  title={Neural discrete representation learning},
  author={Van Den Oord, Aaron and Vinyals, Oriol and others},
  journal={Advances in Neural Information Processing Systems (NeurIPS)},
  year={2017}
}

@inproceedings{lee2022autoregressive,
  title={Autoregressive image generation using residual quantization},
  author={Lee, Doyup and Kim, Chiheon and Kim, Saehoon and Cho, Minsu and Han, Wook-Shin},
  booktitle={Proceedings of the IEEE/CVF Conference on Computer Vision and Pattern Recognition (CVPR)},
  year={2022}
}

@article{xu2025pisces,
  title={Pisces: An Auto-regressive Foundation Model for Image Understanding and Generation},
  author={Xu, Zhiyang and Chen, Jiuhai and Lin, Zhaojiang and Pan, Xichen and Huang, Lifu and Zhou, Tianyi and Khabsa, Madian and Wang, Qifan and Jin, Di and Yasunaga, Michihiro and others},
  journal={arXiv preprint arXiv:2506.10395},
  year={2025}
}

@article{pan2025transfer,
  title={Transfer between modalities with metaqueries},
  author={Pan, Xichen and Shukla, Satya Narayan and Singh, Aashu and Zhao, Zhuokai and Mishra, Shlok Kumar and Wang, Jialiang and Xu, Zhiyang and Chen, Jiuhai and Li, Kunpeng and Juefei-Xu, Felix and others},
  journal={arXiv preprint arXiv:2504.06256},
  year={2025}
}

@inproceedings{ma2025janusflow,
  title={Janusflow: Harmonizing autoregression and rectified flow for unified multimodal understanding and generation},
  author={Ma, Yiyang and Liu, Xingchao and Chen, Xiaokang and Liu, Wen and Wu, Chengyue and Wu, Zhiyu and Pan, Zizheng and Xie, Zhenda and Zhang, Haowei and Yu, Xingkai and others},
  booktitle={Proceedings of the Computer Vision and Pattern Recognition Conference},
  pages={7739--7751},
  year={2025}
}

@inproceedings{wu2025janus,
  title={Janus: Decoupling visual encoding for unified multimodal understanding and generation},
  author={Wu, Chengyue and Chen, Xiaokang and Wu, Zhiyu and Ma, Yiyang and Liu, Xingchao and Pan, Zizheng and Liu, Wen and Xie, Zhenda and Yu, Xingkai and Ruan, Chong and others},
  booktitle={Proceedings of the Computer Vision and Pattern Recognition Conference},
  pages={12966--12977},
  year={2025}
}

@article{han2025vision,
  title={Vision as a Dialect: Unifying Visual Understanding and Generation via Text-Aligned Representations},
  author={Han, Jiaming and Chen, Hao and Zhao, Yang and Wang, Hanyu and Zhao, Qi and Yang, Ziyan and He, Hao and Yue, Xiangyu and Jiang, Lu},
  journal={arXiv preprint arXiv:2506.18898},
  year={2025}
}

@article{chen2025blip3o,
  title={Blip3o-next: Next frontier of native image generation},
  author={Chen, Jiuhai and Xue, Le and Xu, Zhiyang and Pan, Xichen and Yang, Shusheng and Qin, Can and Yan, An and Zhou, Honglu and Chen, Zeyuan and Huang, Lifu and others},
  journal={arXiv preprint arXiv:2510.15857},
  year={2025}
}

@article{sun2024autoregressive,
  title={Autoregressive model beats diffusion: Llama for scalable image generation},
  author={Sun, Peize and Jiang, Yi and Chen, Shoufa and Zhang, Shilong and Peng, Bingyue and Luo, Ping and Yuan, Zehuan},
  journal={arXiv preprint arXiv:2406.06525},
  year={2024}
}

@article{tong2026scaling,
  title={Scaling Text-to-Image Diffusion Transformers with Representation Autoencoders},
  author={Tong, Shengbang and Zheng, Boyang and Wang, Ziteng and Tang, Bingda and Ma, Nanye and Brown, Ellis and Yang, Jihan and Fergus, Rob and LeCun, Yann and Xie, Saining},
  journal={arXiv preprint arXiv:2601.16208},
  year={2026}
}

@article{ma2025unitok,
  title={Unitok: A unified tokenizer for visual generation and understanding},
  author={Ma, Chuofan and Jiang, Yi and Wu, Junfeng and Yang, Jihan and Yu, Xin and Yuan, Zehuan and Peng, Bingyue and Qi, Xiaojuan},
  journal={arXiv preprint arXiv:2502.20321},
  year={2025}
}

@article{geng2025xomnireinforcementlearningmakes,
  title={X-omni: Reinforcement learning makes discrete autoregressive image generative models great again},
  author={Geng, Zigang and Wang, Yibing and Ma, Yeyao and Li, Chen and Rao, Yongming and Gu, Shuyang and Zhong, Zhao and Lu, Qinglin and Hu, Han and Zhang, Xiaosong and others},
  journal={arXiv preprint arXiv:2507.22058},
  year={2025}
}

@article{wang2024emu3,
  title={Emu3: Next-token prediction is all you need},
  author={Wang, Xinlong and Zhang, Xiaosong and Luo, Zhengxiong and Sun, Quan and Cui, Yufeng and Wang, Jinsheng and Zhang, Fan and Wang, Yueze and Li, Zhen and Yu, Qiying and others},
  journal={arXiv preprint arXiv:2409.18869},
  year={2024}
}

@article{yang2024cogvideox,
  title={Cogvideox: Text-to-video diffusion models with an expert transformer},
  author={Yang, Zhuoyi and Teng, Jiayan and Zheng, Wendi and Ding, Ming and Huang, Shiyu and Xu, Jiazheng and Yang, Yuanming and Hong, Wenyi and Zhang, Xiaohan and Feng, Guanyu and others},
  journal={arXiv preprint arXiv:2408.06072},
  year={2024}
}

@inproceedings{radford2021learning,
  title={Learning transferable visual models from natural language supervision},
  author={Radford, Alec and Kim, Jong Wook and Hallacy, Chris and Ramesh, Aditya and Goh, Gabriel and Agarwal, Sandhini and Sastry, Girish and Askell, Amanda and Mishkin, Pamela and Clark, Jack and others},
  booktitle={International Conference on Machine Learning (ICML)},
  year={2021},
}

@inproceedings{sun2024generative,
  title={Generative multimodal models are in-context learners},
  author={Sun, Quan and Cui, Yufeng and Zhang, Xiaosong and Zhang, Fan and Yu, Qiying and Wang, Yueze and Rao, Yongming and Liu, Jingjing and Huang, Tiejun and Wang, Xinlong},
  booktitle={Proceedings of the IEEE/CVF Conference on Computer Vision and Pattern Recognition (CVPR)},
  year={2024}
}

@article{li2025manzano,
  title={MANZANO: A Simple and Scalable Unified Multimodal Model with a Hybrid Vision Tokenizer},
  author={Li, Yanghao and Qian, Rui and Pan, Bowen and Zhang, Haotian and Huang, Haoshuo and Zhang, Bowen and Tong, Jialing and You, Haoxuan and Du, Xianzhi and Gan, Zhe and others},
  journal={arXiv preprint arXiv:2509.16197},
  year={2025}
}

@inproceedings{rombach2022high,
  title={High-resolution image synthesis with latent diffusion models},
  author={Rombach, Robin and Blattmann, Andreas and Lorenz, Dominik and Esser, Patrick and Ommer, Bj{\"o}rn},
  booktitle={Proceedings of the IEEE/CVF Conference on Computer Vision and Pattern Recognition (CVPR)},
  year={2022}
}

@article{kim2025democratizing,
  title={Democratizing text-to-image masked generative models with compact text-aware one-dimensional tokens},
  author={Kim, Dongwon and He, Ju and Yu, Qihang and Yang, Chenglin and Shen, Xiaohui and Kwak, Suha and Chen, Liang-Chieh},
  journal={arXiv preprint arXiv:2501.07730},
  year={2025}
}

@article{xie2024sana,
  title={Sana: Efficient high-resolution image synthesis with linear diffusion transformers},
  author={Xie, Enze and Chen, Junsong and Chen, Junyu and Cai, Han and Tang, Haotian and Lin, Yujun and Zhang, Zhekai and Li, Muyang and Zhu, Ligeng and Lu, Yao and others},
  journal={arXiv preprint arXiv:2410.10629},
  year={2024}
}

@inproceedings{zhu2025addressing,
  title={Addressing representation collapse in vector quantized models with one linear layer},
  author={Zhu, Yongxin and Li, Bocheng and Xin, Yifei and Xia, Zhihua and Xu, Linli},
  booktitle={Proceedings of the IEEE/CVF International Conference on Computer Vision (ICCV)},
  year={2025}
}

@article{qian2025pico,
  title={Pico-Banana-400K: A Large-Scale Dataset for Text-Guided Image Editing},
  author={Qian, Yusu and Bocek-Rivele, Eli and Song, Liangchen and Tong, Jialing and Yang, Yinfei and Lu, Jiasen and Hu, Wenze and Gan, Zhe},
  journal={arXiv preprint arXiv:2510.19808},
  year={2025}
}

@article{deng2025emerging,
  title={Emerging properties in unified multimodal pretraining},
  author={Deng, Chaorui and Zhu, Deyao and Li, Kunchang and Gou, Chenhui and Li, Feng and Wang, Zeyu and Zhong, Shu and Yu, Weihao and Nie, Xiaonan and Song, Ziang and others},
  journal={arXiv preprint arXiv:2505.14683},
  year={2025}
}

@article{zhang2026nextflow,
  title={NextFlow: Unified Sequential Modeling Activates Multimodal Understanding and Generation},
  author={Zhang, Huichao and Qu, Liao and Liu, Yiheng and Chen, Hang and Song, Yangyang and Dong, Yongsheng and Sun, Shikun and Li, Xian and Wang, Xu and Jiang, Yi and others},
  journal={arXiv preprint arXiv:2601.02204},
  year={2026}
}

@article{chen2025unihetero,
  title={UniHetero: Could Generation Enhance Understanding for Vision-Language-Model at Large Data Scale?},
  author={Chen, Fengjiao and Jing, Minhao and Lu, Weitao and Feng, Yan and Li, Xiaoyu and Cao, Xuezhi},
  journal={arXiv preprint arXiv:2512.23512},
  year={2025}
}

@misc{kakaobrain2022coyo-700m,
  title         = {COYO-700M: Image-Text Pair Dataset},
  author        = {Byeon, Minwoo and Park, Beomhee and Kim, Haecheon and Lee, Sungjun and Baek, Woonhyuk and Kim, Saehoon},
  year          = {2022},
  howpublished  = {\url{https://github.com/kakaobrain/coyo-dataset}},
}

@article{gadre2023datacomp,
  title={Datacomp: In search of the next generation of multimodal datasets},
  author={Gadre, Samir Yitzhak and Ilharco, Gabriel and Fang, Alex and Hayase, Jonathan and Smyrnis, Georgios and Nguyen, Thao and Marten, Ryan and Wortsman, Mitchell and Ghosh, Dhruba and Zhang, Jieyu and others},
  journal={Advances in Neural Information Processing Systems (NeurIPS)},
  year={2023}
}

@article{schuhmann2022laion,
  title={Laion-5b: An open large-scale dataset for training next generation image-text models},
  author={Schuhmann, Christoph and Beaumont, Romain and Vencu, Richard and Gordon, Cade and Wightman, Ross and Cherti, Mehdi and Coombes, Theo and Katta, Aarush and Mullis, Clayton and Wortsman, Mitchell and others},
  journal={Advances in Neural Information Processing Systems (NeurIPS)},
  year={2022}
}

@article{gu2024infinity,
  title={Infinity-mm: Scaling multimodal performance with large-scale and high-quality instruction data},
  author={Gu, Shuhao and Zhang, Jialing and Zhou, Siyuan and Yu, Kevin and Xing, Zhaohu and Wang, Liangdong and Cao, Zhou and Jia, Jintao and Zhang, Zhuoyi and Wang, Yixuan and others},
  journal={arXiv preprint arXiv:2410.18558},
  year={2024}
}

@inproceedings{changpinyo2021conceptual,
  title={Conceptual 12m: Pushing web-scale image-text pre-training to recognize long-tail visual concepts},
  author={Changpinyo, Soravit and Sharma, Piyush and Ding, Nan and Soricut, Radu},
  booktitle={Proceedings of the IEEE/CVF Conference on Computer Vision and Pattern Recognition (CVPR)},
  year={2021}
}

@article{meyer2024public,
  title={Public domain 12m: A highly aesthetic image-text dataset with novel governance mechanisms},
  author={Meyer, Jordan and Padgett, Nick and Miller, Cullen and Exline, Laura},
  journal={arXiv preprint arXiv:2410.23144},
  year={2024}
}

@inproceedings{sharma2018conceptual,
  title = {Conceptual Captions: A Cleaned, Hypernymed, Image Alt-text Dataset For Automatic Image Captioning},
  author = {Sharma, Piyush and Ding, Nan and Goodman, Sebastian and Soricut, Radu},
  booktitle = {Proceedings of ACL},
  year = {2018},
}

@article{li2024DenseFusion,
      title={DenseFusion-1M: Merging Vision Experts for Comprehensive Multimodal Perception}, 
      author={Xiaotong Li and Fan Zhang and Haiwen Diao and Yueze Wang and Xinlong Wang and Ling-Yu Duan},
      year={2024},
      journal={2407.08303},
}

@misc{cui2025comprehensive,
            author       = {Erfei Cui and Yinan He and Zheng Ma and Zhe Chen and Hao Tian and Weiyun Wang and Kunchang Li and Yi Wang and Wenhai Wang and Xizhou Zhu and Lewei Lu and Tong Lu and Yali Wang and Limin Wang and Yu Qiao and Jifeng Dai},
            title        = {ShareGPT-4o: Comprehensive Multimodal Annotations With GPT-4o},
            year         = {2024},
            url          = {https://sharegpt4o.github.io/}, 
}

@article{laurencon2024building,
  title={Building and better understanding vision-language models: insights and future directions},
  author={Lauren{\c{c}}on, Hugo and Marafioti, Andr{\'e}s and Sanh, Victor and Tronchon, L{\'e}o},
  journal={arXiv preprint arXiv:2408.12637},
  year={2024}
}

@article{li2024llavanext-interleave,
  title={Llava-next-interleave: Tackling multi-image, video, and 3d in large multimodal models},
  author={Li, Feng and Zhang, Renrui and Zhang, Hao and Zhang, Yuanhan and Li, Bo and Li, Wei and Ma, Zejun and Li, Chunyuan},
  journal={arXiv preprint arXiv:2407.07895},
  year={2024}
}

@inproceedings{chen2024sharegpt4v,
            title={Sharegpt4v: Improving large multi-modal models with better captions},
            author={Chen, Lin and Li, Jinsong and Dong, Xiaoyi and Zhang, Pan and He, Conghui and Wang, Jiaqi and Zhao, Feng and Lin, Dahua},
            booktitle={European Conference on Computer Vision (ECCV)},
            year={2024},
          }

@inproceedings{kafle2018dvqa,
  title={DVQA: Understanding Data Visualizations via Question Answering},
  author={Kafle, Kushal and Cohen, Scott and Price, Brian and Kanan, Christopher},
  booktitle={Proceedings of the IEEE/CVF Conference on Computer Vision and Pattern Recognition (CVPR)},
  year={2018}
}

@article{zhang2024mavis,
  title={Mavis: Mathematical visual instruction tuning},
  author={Zhang, Renrui and Wei, Xinyu and Jiang, Dongzhi and Zhang, Yichi and Guo, Ziyu and Tong, Chengzhuo and Liu, Jiaming and Zhou, Aojun and Wei, Bin and Zhang, Shanghang and others},
  journal={arXiv e-prints},
  pages={arXiv--2407},
  year={2024}
}

@article{bansal2025honeybee,
  title={HoneyBee: Data Recipes for Vision-Language Reasoners},
  author={Bansal, Hritik and Sachan, Devandra Singh and Chang, Kai-Wei and Grover, Aditya and Ghosh, Gargi and Yih, Wen-tau and Pasunuru, Ramakanth},
  journal={arXiv preprint arXiv:2510.12225},
  year={2025}
}

@article{wiedmann2025finevision,
  title={Finevision: Open data is all you need},
  author={Wiedmann, Luis and Zohar, Orr and Mahla, Amir and Wang, Xiaohan and Li, Rui and Frere, Thibaud and von Werra, Leandro and Gosthipaty, Aritra Roy and Marafioti, Andr{\'e}s},
  journal={arXiv preprint arXiv:2510.17269},
  year={2025}
}

@article{tian2024mm,
  title={Mm-interleaved: Interleaved image-text generative modeling via multi-modal feature synchronizer},
  author={Tian, Changyao and Zhu, Xizhou and Xiong, Yuwen and Wang, Weiyun and Chen, Zhe and Wang, Wenhai and Chen, Yuntao and Lu, Lewei and Lu, Tong and Zhou, Jie and others},
  journal={arXiv preprint arXiv:2401.10208},
  year={2024}
}

@ONLINE{wikidump,
    author = "Wikimedia Foundation",
    title  = "Wikimedia Downloads",
    url    = "https://dumps.wikimedia.org"
}

@ONLINE{grok,
    author = "Grok",
    title  = "Grok",
    url    = "https://x. ai/blog/grok-1.5"
}

@inproceedings{gokaslan2024commoncanvas,
  title={Commoncanvas: Open diffusion models trained on creative-commons images},
  author={Gokaslan, Aaron and Cooper, A Feder and Collins, Jasmine and Seguin, Landan and Jacobson, Austin and Patel, Mihir and Frankle, Jonathan and Stephenson, Cory and Kuleshov, Volodymyr},
  booktitle={Proceedings of the IEEE/CVF Conference on Computer Vision and Pattern Recognition (CVPR)},
  year={2024}
}

@misc{midjourney-niji-1m-llavanext,
  author = { Caption Emporium },
  title = { midjourney-niji-1m-llavanext },
  year = { 2024 },
  publisher = { Huggingface },
  journal = { Huggingface repository },
  howpublished = {\url{https://huggingface.co/datasets/CaptionEmporium/conceptual-captions-cc12m-llavanext}},
}

@article{chen2025blip3,
  title={Blip3-o: A family of fully open unified multimodal models-architecture, training and dataset},
  author={Chen, Jiuhai and Xu, Zhiyang and Pan, Xichen and Hu, Yushi and Qin, Can and Goldstein, Tom and Huang, Lifu and Zhou, Tianyi and Xie, Saining and Savarese, Silvio and others},
  journal={arXiv preprint arXiv:2505.09568},
  year={2025}
}

@article{team2025kwai,
  title={Kwai Keye-VL Technical Report},
  author={Team, Kwai Keye and Yang, Biao and Wen, Bin and Liu, Changyi and Chu, Chenglong and Song, Chengru and Rao, Chongling and Yi, Chuan and Li, Da and Zang, Dunju and others},
  journal={arXiv preprint arXiv:2507.01949},
  year={2025}
}

@article{alayrac2022flamingo,
  title={Flamingo: a visual language model for few-shot learning},
  author={Alayrac, Jean-Baptiste and Donahue, Jeff and Luc, Pauline and Miech, Antoine and Barr, Iain and Hasson, Yana and Lenc, Karel and Mensch, Arthur and Millican, Katherine and Reynolds, Malcolm and others},
  journal={Advances in Neural Information Processing Systems (NeurIPS)},
  year={2022}
}

@inproceedings{li2023blip,
  title={Blip-2: Bootstrapping language-image pre-training with frozen image encoders and large language models},
  author={Li, Junnan and Li, Dongxu and Savarese, Silvio and Hoi, Steven},
  booktitle={International Conference on Machine Learning (ICML)},
  year={2023},
}

@inproceedings{liu2024mmbench,
  title={Mmbench: Is your multi-modal model an all-around player?},
  author={Liu, Yuan and Duan, Haodong and Zhang, Yuanhan and Li, Bo and Zhang, Songyang and Zhao, Wangbo and Yuan, Yike and Wang, Jiaqi and He, Conghui and Liu, Ziwei and others},
  booktitle={European Conference on Computer Vision (ECCV)},
  year={2024},
}

@inproceedings{kembhavi2016diagram,
  title={A diagram is worth a dozen images},
  author={Kembhavi, Aniruddha and Salvato, Mike and Kolve, Eric and Seo, Minjoon and Hajishirzi, Hannaneh and Farhadi, Ali},
  booktitle={European Conference on Computer Vision (ECCV)},
  year={2016},
}

@inproceedings{yue2024mmmu,
  title={Mmmu: A massive multi-discipline multimodal understanding and reasoning benchmark for expert agi},
  author={Yue, Xiang and Ni, Yuansheng and Zhang, Kai and Zheng, Tianyu and Liu, Ruoqi and Zhang, Ge and Stevens, Samuel and Jiang, Dongfu and Ren, Weiming and Sun, Yuxuan and others},
  booktitle={Proceedings of the IEEE/CVF Conference on Computer Vision and Pattern Recognition (CVPR)},
  year={2024}
}

@article{lu2023mathvista,
  title={Mathvista: Evaluating mathematical reasoning of foundation models in visual contexts},
  author={Lu, Pan and Bansal, Hritik and Xia, Tony and Liu, Jiacheng and Li, Chunyuan and Hajishirzi, Hannaneh and Cheng, Hao and Chang, Kai-Wei and Galley, Michel and Gao, Jianfeng},
  journal={arXiv preprint arXiv:2310.02255},
  year={2023}
}

@article{liu2024chatqa,
  title={ChatQA: Surpassing GPT-4 on Conversational QA and RAG},
  author={Liu, Zihan and Ping, Wei and Roy, Rajarshi and Xu, Peng and Lee, Chankyu and Shoeybi, Mohammad and Catanzaro, Bryan},
  journal={arXiv preprint arXiv:2401.10225},
  year={2024}}

@inproceedings{singh2019towards,
    title={Towards VQA Models That Can Read},
    author={Singh, Amanpreet and Natarjan, Vivek and Shah, Meet and Jiang, Yu and Chen, Xinlei and Batra, Dhruv and Parikh, Devi and Rohrbach, Marcus},
    booktitle={Proceedings of the IEEE/CVF Conference on Computer Vision and Pattern Recognition (CVPR)},
    year={2019}
}

@article{Liu_2024,
  title={Ocrbench: on the hidden mystery of ocr in large multimodal models},
  author={Liu, Yuliang and Li, Zhang and Huang, Mingxin and Yang, Biao and Yu, Wenwen and Li, Chunyuan and Yin, Xu-Cheng and Liu, Cheng-Lin and Jin, Lianwen and Bai, Xiang},
  journal={Science China Information Sciences},
  volume={67},
  number={12},
  pages={220102},
  year={2024}
}

@article{ghosh2023geneval,
  title={Geneval: An object-focused framework for evaluating text-to-image alignment},
  author={Ghosh, Dhruba and Hajishirzi, Hannaneh and Schmidt, Ludwig},
  journal={Advances in Neural Information Processing Systems (NeurIPS)},
  year={2023}
}

@article{niu2025wise,
  title={Wise: A world knowledge-informed semantic evaluation for text-to-image generation},
  author={Niu, Yuwei and Ning, Munan and Zheng, Mengren and Jin, Weiyang and Lin, Bin and Jin, Peng and Liao, Jiaqi and Feng, Chaoran and Ning, Kunpeng and Zhu, Bin and others},
  journal={arXiv preprint arXiv:2503.07265},
  year={2025}
}

@article{chen2025blip3ofamilyfullyopen,
  title={Blip3-o: A family of fully open unified multimodal models-architecture, training and dataset},
  author={Chen, Jiuhai and Xu, Zhiyang and Pan, Xichen and Hu, Yushi and Qin, Can and Goldstein, Tom and Huang, Lifu and Zhou, Tianyi and Xie, Saining and Savarese, Silvio and others},
  journal={arXiv preprint arXiv:2505.09568},
  year={2025}
}

@article{mingUniVision,
  title={Ming-univision: Joint image understanding and generation with a unified continuous tokenizer},
  author={Huang, Ziyuan and Zheng, DanDan and Zou, Cheng and Liu, Rui and Wang, Xiaolong and Ji, Kaixiang and Chai, Weilong and Sun, Jianxin and Wang, Libin and Lv, Yongjie and others},
  journal={arXiv preprint arXiv:2510.06590},
  year={2025}
}

@article{wang2024qwen2,
  title={Qwen2-vl: Enhancing vision-language model's perception of the world at any resolution},
  author={Wang, Peng and Bai, Shuai and Tan, Sinan and Wang, Shijie and Fan, Zhihao and Bai, Jinze and Chen, Keqin and Liu, Xuejing and Wang, Jialin and Ge, Wenbin and others},
  journal={arXiv preprint arXiv:2409.12191},
  year={2024}
}

@inproceedings{chen2024internvl,
  title={Internvl: Scaling up vision foundation models and aligning for generic visual-linguistic tasks},
  author={Chen, Zhe and Wu, Jiannan and Wang, Wenhai and Su, Weijie and Chen, Guo and Xing, Sen and Zhong, Muyan and Zhang, Qinglong and Zhu, Xizhou and Lu, Lewei and others},
  booktitle={Proceedings of the IEEE/CVF Conference on Computer Vision and Pattern Recognition (CVPR)},
  year={2024}
}

@article{bai2025qwen2,
  title={Qwen2. 5-vl technical report},
  author={Bai, Shuai and Chen, Keqin and Liu, Xuejing and Wang, Jialin and Ge, Wenbin and Song, Sibo and Dang, Kai and Wang, Peng and Wang, Shijie and Tang, Jun and others},
  journal={arXiv preprint arXiv:2502.13923},
  year={2025}
}

@article{guo2025seed1,
  title={Seed1. 5-vl technical report},
  author={Guo, Dong and Wu, Faming and Zhu, Feida and Leng, Fuxing and Shi, Guang and Chen, Haobin and Fan, Haoqi and Wang, Jian and Jiang, Jianyu and Wang, Jiawei and others},
  journal={arXiv preprint arXiv:2505.07062},
  year={2025}
}

@article{team2024chameleon,
  title={Chameleon: Mixed-modal early-fusion foundation models},
  author={Team, Chameleon},
  journal={arXiv preprint arXiv:2405.09818},
  year={2024}
}

@article{liu2025tuna,
  title={TUNA: Taming Unified Visual Representations for Native Unified Multimodal Models},
  author={Liu, Zhiheng and Ren, Weiming and Liu, Haozhe and Zhou, Zijian and Chen, Shoufa and Qiu, Haonan and Huang, Xiaoke and An, Zhaochong and Yang, Fanny and Patel, Aditya and others},
  journal={arXiv preprint arXiv:2512.02014},
  year={2025}
}

@article{team2025zimage,
  title={Z-Image: An Efficient Image Generation Foundation Model with Single-Stream Diffusion Transformer},
  author={Z-Image Team},
  journal={arXiv preprint arXiv:2511.22699},
  year={2025}
}

@article{xie2025sana,
  title={Sana 1.5: Efficient scaling of training-time and inference-time compute in linear diffusion transformer},
  author={Xie, Enze and Chen, Junsong and Zhao, Yuyang and Yu, Jincheng and Zhu, Ligeng and Wu, Chengyue and Lin, Yujun and Zhang, Zhekai and Li, Muyang and Chen, Junyu and others},
  journal={arXiv preprint arXiv:2501.18427},
  year={2025}
}

@article{song2025dualtoken,
  title={Dualtoken: Towards unifying visual understanding and generation with dual visual vocabularies},
  author={Song, Wei and Wang, Yuran and Song, Zijia and Li, Yadong and Sun, Haoze and Chen, Weipeng and Zhou, Zenan and Xu, Jianhua and Wang, Jiaqi and Yu, Kaicheng},
  journal={arXiv preprint arXiv:2503.14324},
  year={2025}
}

@article{zheng2025onevision,
  title={OneVision: An End-to-End Generative Framework for Multi-view E-commerce Vision Search},
  author={Zheng, Zexin and Dai, Huangyu and Mao, Lingtao and Sun, Xinyu and Liang, Zihan and Chen, Ben and Ding, Yuqing and Lei, Chenyi and Ou, Wenwu and Li, Han and others},
  journal={arXiv preprint arXiv:2510.05759},
  year={2025}
}

@inproceedings{kim2022ocr,
  title={Ocr-free document understanding transformer},
  author={Kim, Geewook and Hong, Teakgyu and Yim, Moonbin and Nam, JeongYeon and Park, Jinyoung and Yim, Jinyeong and Hwang, Wonseok and Yun, Sangdoo and Han, Dongyoon and Park, Seunghyun},
  booktitle={European Conference on Computer Vision},
  pages={498--517},
  year={2022},
  organization={Springer}
}

@inproceedings{xu2018towards,
  title={Towards End-to-End License Plate Detection and Recognition: A Large Dataset and Baseline},
  author={Xu, Zhenbo and Yang, Wei and Meng, Ajin and Lu, Nanxue and Huang, Huan},
  booktitle={Proceedings of the European Conference on Computer Vision (ECCV)},
  pages={255--271},
  year={2018}
}

@article{codeocr_leetcode_2025,
  author       = {Maksonchek},
  title        = {CodeOCR Dataset (Python Code Images + Ground Truth)},
  year         = {2025},
  publisher    = {Hugging Face},
  url          = {https://huggingface.co/datasets/maksonchek/codeocr-dataset}
}

@inproceedings{duan2025instructocr,
  title={InstructOCR: Instruction Boosting Scene Text Spotting},
  author={Duan, Chen and Jiang, Qianyi and Fu, Pei and Chen, Jiamin and Li, Shengxi and Wang, Zining and Guo, Shan and Luo, Junfeng},
  booktitle={Proceedings of the AAAI Conference on Artificial Intelligence},
  volume={39},
  number={3},
  pages={2807--2815},
  year={2025}
}

@misc{llamagen,
      title={Autoregressive Model Beats Diffusion: Llama for Scalable Image Generation}, 
      author={Peize Sun and Yi Jiang and Shoufa Chen and Shilong Zhang and Bingyue Peng and Ping Luo and Zehuan Yuan},
      year={2024},
      eprint={2406.06525},
      archivePrefix={arXiv},
      primaryClass={cs.CV},
      url={https://arxiv.org/abs/2406.06525}, 
}

@article{jegou2011product,
  title={Product quantization for nearest neighbor search},
  author={J{\'e}gou, Herv{\'e} and Douze, Matthijs and Schmid, Cordelia},
  journal={IEEE Transactions on Pattern Analysis and Machine Intelligence},
  volume={33},
  number={1},
  pages={117--128},
  year={2011}
}
\bibliographystyle{IEEEtranN}

\appendix

\begin{figure}[t!]
  \centering
  \includegraphics[width=0.95\textwidth]{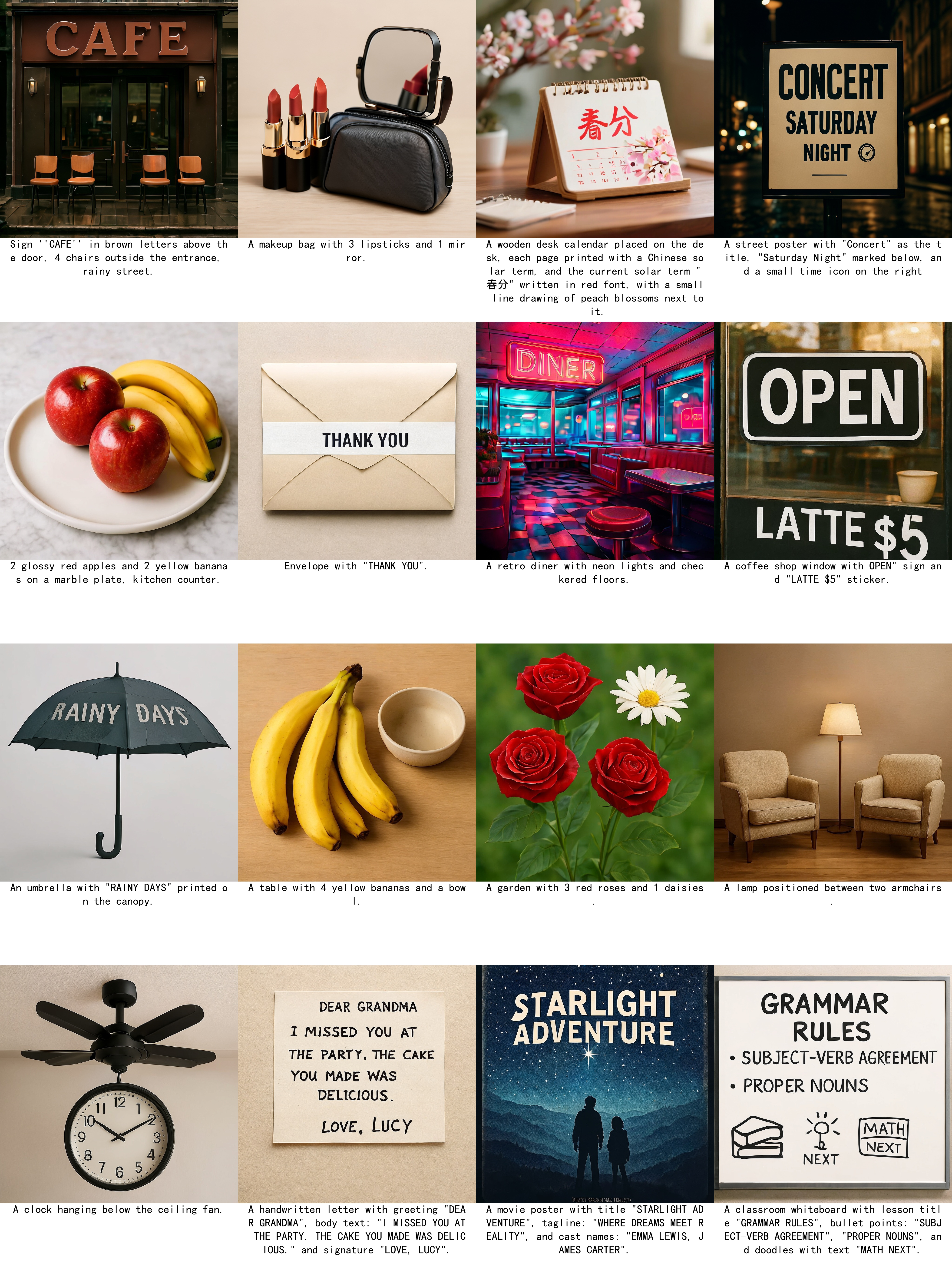}
    \caption{The visualization of Kelix in image generation.}
    \label{fig:showcase}
\end{figure}

\begin{figure}[t!]
  \centering
  \includegraphics[width=0.95\textwidth]{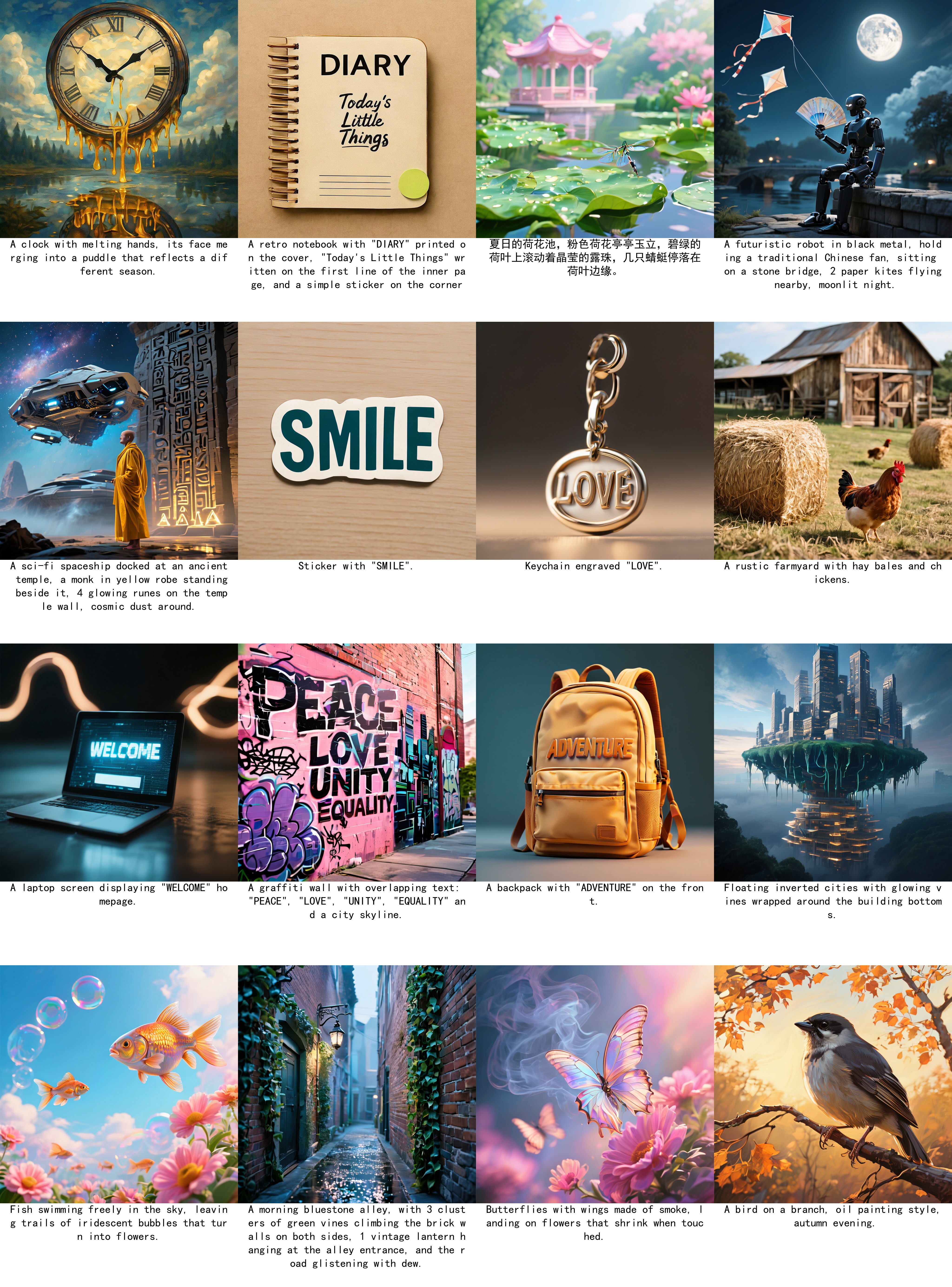}
    \caption{The visualization of Kelix in image generation.}
    \label{fig:showcase}
\end{figure}

\begin{figure}[t!]
  \centering
  \includegraphics[width=0.95\textwidth]{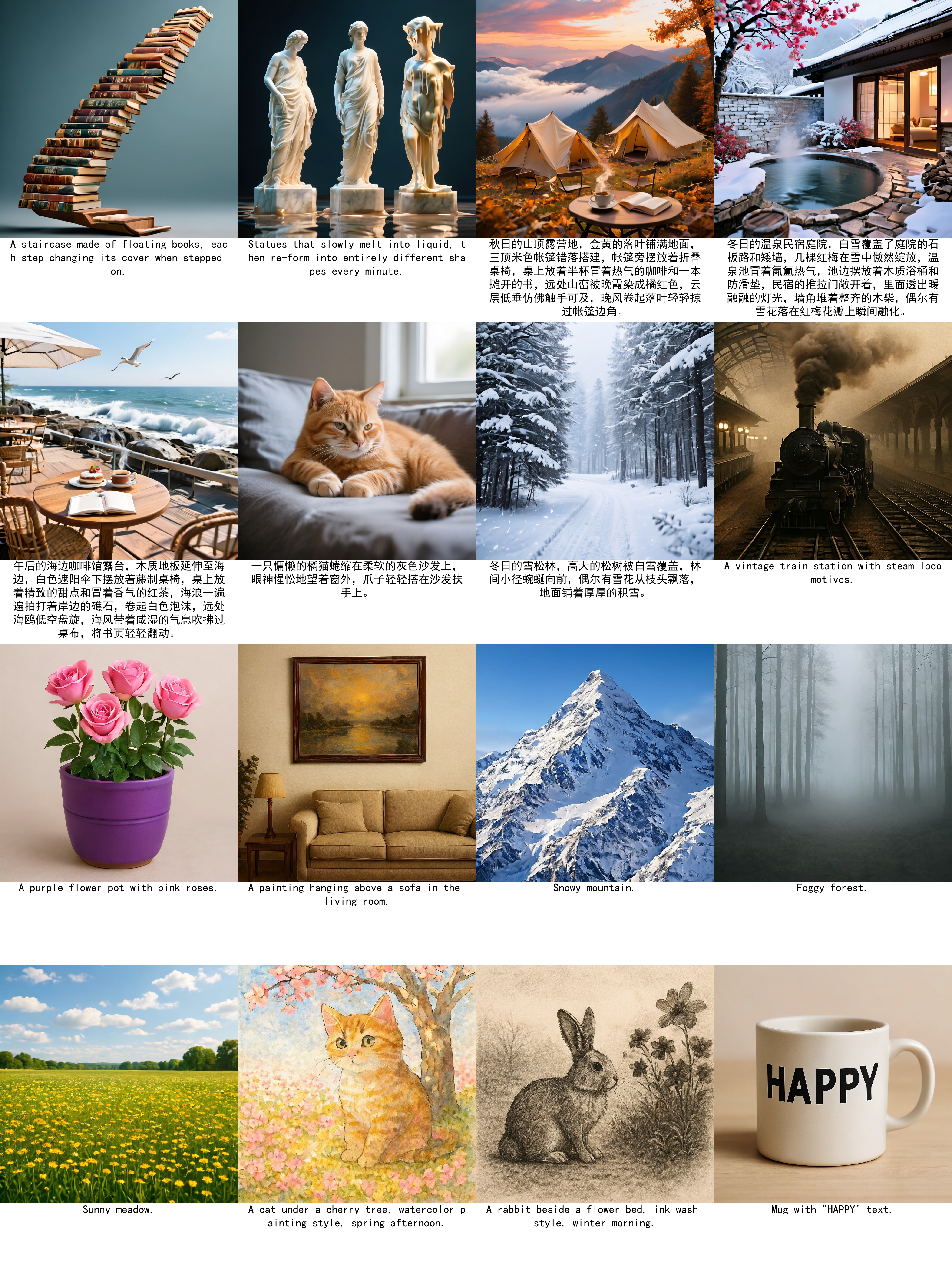}
    \caption{The visualization of Kelix in image generation.}
    \label{fig:showcase}
\end{figure}

\newpage
\quad \\
\quad \\

\section{Contributions}

\noindent
\textbf{Core Contributors}\quad
Boyang Ding\textsuperscript{*},
Chenglong Chu,
Dunju Zang,
Han Li,
Jiangxia Cao,
Kun Gai,
Muhao Wei,
Ruiming Tang,
Shiyao Wang,
Siyang Mao,
Xinchen Luo,
Yahui Liu,
Zhixin Ling,
Zhuoran Yang,
Ziming Li


\vspace{0.5em}
\noindent
\textbf{Contributors}\quad
Chengru Song,
Guorui Zhou,
Guowang Zhang,
Hao Peng,
Hao Wang,
Jiaxin Deng,
Jin Ouyang,
Jinghao Zhang,
Lejian Ren,
Qianqian Wang,
Qigen Hu,
Tao Wang,
Xingmei Wang,
Yiping Yang,
Zixing Zhang,
Ziqi Wang

\vspace{0.5em}
\noindent
{\small All the authors listed alphabetically by first name.\quad \textsuperscript{*}Project Leader.}

\end{document}